\documentclass{article}

    \PassOptionsToPackage{numbers, compress, square}{natbib}

\usepackage[preprint]{neurips_data_2023}

\usepackage[utf8]{inputenc} 
\usepackage[T1]{fontenc}    
\usepackage{hyperref}       
\usepackage{url}            
\usepackage{booktabs}       
\usepackage{amsfonts}       
\usepackage{nicefrac}       
\usepackage{microtype}      
\usepackage{xcolor}         
\usepackage[colorinlistoftodos]{todonotes} 
\usepackage{enumitem}
\usepackage{scalerel}
\usepackage{soul}
\sethlcolor{green}

\usepackage{amsmath}
\usepackage{multirow}

\usepackage{graphics}
\graphicspath{{/figs}}
\usepackage{tikz}
\usepackage{comment}
\usepackage{subcaption}
\usepackage{floatrow}
\usepackage{wrapfig}
\usepackage{listings}
\newfloatcommand{capbtabbox}{table}[][\FBwidth]
\newcommand\R{{\mathbb R}}

\newcommand{\greencheck}{}%
\DeclareRobustCommand{\greencheck}{%
  \tikz\fill[scale=0.4, color=green]
  (0,.35) -- (.25,0) -- (1,.7) -- (.25,.15) -- cycle;%
}

\newcommand{\blackcheck}{}%
\DeclareRobustCommand{\blackcheck}{%
  \tikz\fill[scale=0.4, color=black]
  (0,.35) -- (.25,0) -- (1,.7) -- (.25,.15) -- cycle;%
}

\title{Routing Arena: A Benchmark Suite for Neural Routing Solvers}

\author{
   Daniela Thyssens \\
   Information Systems and ML Lab, \\
   University of Hildesheim, Germany \\
   \texttt{thyssens@ismll.uni-hildesheim.de} \\
   \And
   Tim Dernedde \\
   Information Systems and ML Lab, \\ 
   University of Hildesheim, Germany \\
   \texttt{dernedde@ismll.uni-hildesheim.de} \\
  \And
   Jonas Falkner \\
   Information Systems and ML Lab, \\
   University of Hildesheim, Germany \\
   \texttt{falkner@ismll.uni-hildesheim.de} \\
      \And
   Lars Schmidt-Thieme \\
   Information Systems and ML Lab, \\ 
   University of Hildesheim, Germany \\
   \texttt{schmidt-thieme@ismll.uni-hildesheim.de} \\
}

\begin{document}

\maketitle


\begin{abstract}
  Neural Combinatorial Optimization has been researched actively in the last eight years. Even though many of the proposed Machine Learning based approaches are compared on the same datasets, the evaluation protocol exhibits essential flaws and the selection of baselines often neglects State-of-the-Art Operations Research approaches. To improve on both of these shortcomings, we propose the Routing Arena, a benchmark suite for Routing Problems that provides a seamless integration of consistent evaluation and the provision of baselines and benchmarks prevalent in the Machine Learning- and Operations Research field. The proposed evaluation protocol considers the two most important evaluation cases for different applications: First, the solution quality for an a priori fixed time budget and secondly the anytime performance of the respective methods. 
  By setting the solution trajectory in perspective to a Best Known Solution and a Base Solver's solutions trajectory, we furthermore propose the Weighted Relative Average Performance (WRAP), a novel evaluation metric that quantifies the often claimed runtime efficiency of Neural Routing Solvers.
  A comprehensive first experimental evaluation demonstrates that the most recent Operations Research solvers generate state-of-the-art results in terms of solution quality and runtime efficiency when it comes to the vehicle routing problem. Nevertheless, some findings highlight the advantages of neural approaches and motivate a shift in how neural solvers should be conceptualized.
\end{abstract}
\section{Introduction}
\label{intro}
In recent years, the research field of neural combinatorial optimization (NCO) has developed a veritable zoo of methods to solve the vehicle routing problem (VRP) and its variants. The general motivation for NCO is two-fold: 
(i) finding reasonably good solutions faster than traditional approaches
and (ii) saving development effort and hand-crafted engineering by learning a parametrized policy to trade off computational complexity and optimality in solving combinatorial optimization (CO)
problems \cite{kool2018attention}.
The recent surge of established approaches, that often share many architectural and conceptual similarities, has led to a need for a unified evaluation protocol. 
%


So far, the evaluation of runtime efficiency in NCO was based on a method's total runtime, which favors neural approaches, since they often use parallel batch processing to quickly solve large numbers

\begin{wrapfigure}{r}{0.5\textwidth}
    \includegraphics[trim= {1.5cm 0.5cm 1.5cm 0.2cm}]{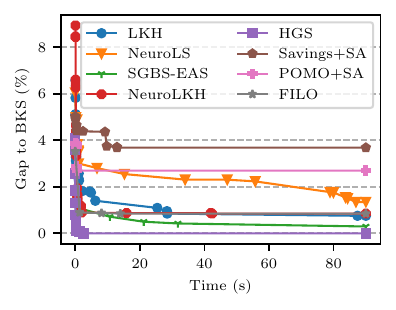} 
  \caption{\% Gap to BKS over the cumulative running time for the first uniformly distributed instance in \cite{kool2018attention} of size 100.}
\label{fig:trajectories_unif100}
\end{wrapfigure}%
of problems on the GPU compared to OR solvers, which operate on a per instance level. Such approaches could also be parallelized over batches, e.g. by running them on different cores of a multi CPU machine, however this comparison has not been considered in the literature so far. Accordingly, the comparison between those methods is not consistent.
Generally, comparing total run-times, even on a per-instance level, to evaluate efficiency is not recommended, as it completely ignores the point of time when the final solution of a local search (LS) has been found along the search trajectory. In fact, the operations research (OR) literature has  developed strategies to improve considerably on the run time complexity issue and brings forth State-of-the-Art meta-heuristics, such as HGS-CVRP (\cite{vidal2022hybrid}), that deliver qualitatively strong solutions within milliseconds (see Figure \ref{fig:trajectories_unif100}).
Thus, to evaluate the overall performance of primal heuristics, i.e. methods that should find good solutions quickly, the solution trajectory over time should be assessed, as has been done in the OR field for years (\cite{berthold2013measuring}). 
Given these two, currently independently evolving, strings of research for routing problems, we advocate the need for a unified evaluation protocol that allows researchers to truthfully assess their methods against State-of-the-Art Neural- as well as OR methods on prevalent benchmarks. Starting with the classic VRP, we propose the \textit{Routing Arena} (RA), a benchmark suite that functions as an assessment center for novel methods and a \textbf{development tool} to promote research. The contribution of the RA can be summarized as follows:
\begin{itemize}
\item \textit{Data \& Model Pool}: The Routing Arena provides an integrated testing suite to benchmark novel methods on over \textbf{15 datasets} against \textbf{nine machine learning- and five OR-based meta-heuristic solvers}. 
\item \textit{The Evaluation}: We propose a \textbf{unified evaluation protocol} together with a \textbf{novel evaluation metric}, Weighted Relative Average Performance (WRAP), to evaluate the performance of a method's solution trajectory, by capturing the relative improvement over a simple baseline as well as its gap to a Best Known Solution (BKS). WRAP aggregates well over instance sets, enabling to better capture the real performance signal across experiments.
\item \textit{Fairness and Integration}: All metric evaluations incorporate hardware information about resources used to perform the benchmark runs in order to neutralize effects on runtime efficiency that stem solely from improved hardware capabilities. Thus, evaluation performed in the benchmark suite produces \textbf{standardized results irrelevant of the machine(s) in use}. 
\item \textit{Diverse Training Pool}: The RA enriches the \textbf{pool of useful training sets} in the L2O community, by integrating data samplers of challenging data distributions. The samplers provide customizable, on the fly data generation for individually configured training runs.
\end{itemize}
\section{Background and Related Work}
\label{related_work}
\paragraph{Motivation for NCO.}
The OR literature has spent decades on identifying relevant combinatorial problems and developing (meta)-heuristics and solvers for them. In order to fruitfully apply data driven methods to such problems, NCO makes the assumption that for an application requiring to solve instances from a given problem class, there exists an underlying distribution over the instances to be solved \cite{kool2018attention, bengioTourdhorizon2021}. Thus, the main motivation stated in the literature for applying ML to CO problems is to learn heuristics or replace parts of heuristics with learned components with the goal of reaching either better overall solutions or high quality solutions faster than traditional handcrafted algorithms \textbf{on the problem distribution} \cite{kool2018attention, kwon2020pomo, eas, JoshiCRL21, li2021learning}. Another motivation is that end-to-end learnable pipelines can be more easily adjusted to new or understudied problem types such that a high quality heuristic can be automatically learned without having to manually engineer a new method \cite{bello2016neural, kool2018attention, JoshiCRL21, kwon2020pomo, choo2022simulation}. This motivation however is harder to evaluate for, since quantifying the difficulty of adjustment for any method to a new problem type is hard. The Routing Arena's main goal is thus to evaluate methods w.r.t. the prior motivation of finding better solutions faster or better overall solutions. In a first instance, the Routing Arena focuses on the capacitated VRP (CVRP), since it is one of the most common problems in NCO and among the most well studied problems in OR, where it is also used as a testbed for new ideas \cite{uchoa2017new}, providing high quality solvers to compare against.
\paragraph{The CVRP.}
%
The CVRP involves a set of customers $C = \{1, ..., n\}$, a depot node $0$, all pairwise distances $d_{ij}, \forall (i,j) \in C \cup 0$, representing the cost of traveling from customer $i$ to $j$, a demand per customer $q_i, i \in C$ and a total vehicle capacity $Q$. A feasible solution $x = \{r_1, ..., r_{|x|}\}$ is a set of routes $r_i = (r_{i,1}, …, r_{i,N_i})$, which is a sequence of customers, always starting and ending at the depot, such that $r_{i,1}=0$ and $r_{i,N_i}=0$ and $N_i$ denotes the length of the route. It is feasible if the cumulative demand of any route does not exceed the vehicle capacity and each customer is visited exactly once. Let $f(x)$ denote the total distance of a solution $x$, then the goal of the CVRP is to find the feasible solution with minimal cost $x^* = \text{argmin}_x f(x)$. 
%
\paragraph{Algorithms for solving the CVRP.}
The scientific community has developed a plethora of (meta)-heuristics and exact algorithms for solving the CVRP over the last 60 years since the problem was first stated in \cite{dantzig1959truck} with a recent surge in learning heuristics or components over a distribution of CVRP instances. It is not in the scope of this work to give an exhaustive overview, instead we refer the reader to surveys \cite{tothVehicleRoutingProblems2014, vrpSurveyMl, bengioTourdhorizon2021, CappartCK00V21}, covering both strings of research. All methods that are currently considered in the RA will be introduced in section \ref{sec:RA} and discussed in more detail in apppendix \ref{app:baselines}.
\paragraph{Existing Benchmarks.} 
The majority of existing datasets and benchmarks for the VRP were contributed from the field of OR. Along the decades of research, many distinct problem instances and small benchmark sets were released together with publications of new approaches. A first notable attempt at gathering different problem instances in one place is the open-data
platform VRP-REP \cite{mendoza2014vrp}, which gathered instances ranging from the \emph{E-Set} (\cite{christofides1969algorithm}) to the set provided by \cite{golden1998impact}. It was updated in \cite{mendoza2017update} and now features the most comprehensive dataset collection of various routing problem variants albeit also in different formats. Another fairly recent collection of standard CVRP Benchmark Datasets is found on CVRPLib \footnote{http://vrp.galgos.inf.puc-rio.br/index.php/en/}, which also constitutes test sets that were used in recent implementation challenges.
Motivated by the fact that many prevailing benchmark sets are lacking discriminating power and are artificially generated, \cite{uchoa2017new} propose a benchmark CVRP set (\emph{X-Set}) that is particularly challenging to solve, while being less artificial and more heterogeneous. The proposed instances are generated on the basis of five attributes (problem size, customer positioning, depot positioning, type of demand distribution and average route size) and constitute the current gold standard for systematically evaluating the performance of solving the CVRP in the OR field. Recently, the \textit{X-Set} appeared increasingly in the NCO field,
which, previously, has predominantly focused its evaluation on the instance set provided in \cite{nazari2018reinforcement}, which consists of problems with uniformly sampled customer-positions in the unit square and uniformly distributed demands for the problem sizes of 20, 50 and 100 and do not come with the documentation of a BKS for each instance, as commonly the solution qualities are averaged or only sub-samples or newly generated samples from the same distribution are compared \cite{xin2021neurolkh}.
Since handling different sized instances within a batch 
often requires not insignificant extra implementation effort, some works (\cite{kool2022deep}, \cite{hottung2019neural}) adapt the benchmark set \emph{X} in \cite{uchoa2017new} by generating smaller equal-sized problem instances from the documented distributions to incorporate this more challenging set in their experimental analysis. 
An experimental evaluation on parts of the original \emph{X} instances for evaluating Neural CVRP approaches is rarely performed (\cite{falkner2023learning}) and mostly discussed in the appendix (\cite{ma2021learning, xin2021neurolkh}).
A summary of which models are currently evaluated on the three most commonly used benchmark datasets of both fields is presented in Table \ref{overview}.
\begin{table}[!ht]

    \centering
    \resizebox{\columnwidth}{!}{
    \renewcommand{\arraystretch}{1.3} 
    \begin{tabular}{llcccccccccccccc}
    \toprule
     &\textbf{Datasets} & \multicolumn{14}{c}{\textbf{Models / Algorithms}} \\
    & & AM & MDAM & POMO & NLNS & SGBS & DPDP & DACT & NLS & NeuroLKH & ORT-GLS & LKH & HGS & FILO \\
    \midrule
    \parbox[t]{5mm}{\multirow{4}{*}{\rotatebox[origin=c]{90}{ML Benchmarks}}} &  & &  &  &  &  &  &  & &  & &  &  &  & \\
    &\citet{nazari2018reinforcement}* & \blackcheck & \blackcheck & \blackcheck & \blackcheck & \blackcheck & \blackcheck & \blackcheck & \blackcheck & \blackcheck & \blackcheck & \blackcheck & \blackcheck & \greencheck \\ 
    &\citet{hottung2019neural} & \greencheck & \greencheck & \greencheck & \blackcheck & \blackcheck & \greencheck & \greencheck & \greencheck & \greencheck & \greencheck& \blackcheck &\greencheck & \greencheck \\
    &\citet{kool2022deep}* & \greencheck & \greencheck & \greencheck & \greencheck & \greencheck & \blackcheck & \greencheck & \greencheck & \greencheck & \greencheck & \blackcheck & \blackcheck & \greencheck\\ 
     \parbox[t]{2mm}{\multirow{5}{*}{\rotatebox[origin=c]{90}{CVRPLib}}} &  & &  &  &  &  &  &  & & &  &  &  & \\
    &\citet{uchoa2017new} & \greencheck & \greencheck & \greencheck & \greencheck & \greencheck &  & \blackcheck & \blackcheck & \blackcheck & \blackcheck & \blackcheck & \blackcheck & \blackcheck\\
    & \citet{golden1998impact} & \greencheck & \greencheck & \greencheck & \greencheck & \greencheck &  & \greencheck & \greencheck & \greencheck & \greencheck & \blackcheck & \blackcheck & \blackcheck \\
    & \citet{queiroga202210}* & \greencheck & \greencheck & \greencheck & \greencheck & \greencheck & \greencheck & \greencheck & \greencheck & \greencheck & \greencheck & \greencheck & \greencheck & \greencheck \\
    
    \bottomrule
    \end{tabular}}
    \caption{ML \& OR Approaches evaluated on existing Datasets. \blackcheck refers to evaluations that have previously been done in the literature. \greencheck are experiments added by us. '*' refers to the instance sets that feature 10000 instances and are evaluated in subsets for benchmark-testing in the Routing Arena.}
    \label{overview}
\end{table}%
The ML methods (first nine columns) for solving the CVRP are mainly evaluated on the uniform \cite{nazari2018reinforcement} Dataset (we focus on the problem size of 100 here), while only the most recent Neural LS approaches are evaluated in parts on the \textit{X} Set. Note that DPDP \cite{kool2022deep} can only be evaluated for instance sets of equal problem size. Concerning the OR based methods we see that only LKH \cite{helsgaun2017extension} and partly HGS-CVRP \cite{vidal2022hybrid} are used as baselines for evaluating on the ML Benchmark Sets.
Furthermore, results on the provided benchmark set by \cite{queiroga202210} have not been published and to the best of our knowledge, this work is the first to benchmark the existing methods in Table \ref{overview} on part of these instances.
\paragraph{Existing ML Evaluation Formats.} 
The typical protocol for evaluating methods on the CVRP in ML consists of two metrics; the objective cost or the gap to a reference solver averaged over a complete dataset and the total time elapsed for solving the complete set. As the main goal of NCO approaches
is to find sufficiently good results in a reasonable amount of time (\cite{falkner2023learning}), the prevailing evaluation protocol accounts for some \textbf{essential flaws}; (1) The \textbf{total runtime} of a method on a set of instances \textbf{is too coarse} and masks how effectively the model solves a given instance. This metric thus mainly evaluates how much parallelization capabilities the algorithm provides in terms of batching and multi-core utilization for the full set of instances, compared to singe instance performance focused by the OR literature. (2) \textbf{Without indication of a per instance time limit} the comparison, especially to OR local search solvers that have a high pre-defined maximum number of iterations, is ill-posed, since they often find their final issued solution within milliseconds for "simple" problem instances. (3) If the goal is to assess efficiency, the \textbf{ML methods should be evaluated on their any-time performance}.
%
%
To improve on the above flaws in evaluation, the Routing Arena proposes to have two protocols attributed to different problem settings for CO problems; One that correctly compares solution quality performances and another to address a method's effectiveness, both of which will be discussed in section \ref{evaluation}.
\section{Routing Arena - Benchmarks, Baselines and Data Generators}
\label{sec:RA}
This section discusses the first contribution highlighted in section \ref{intro}. We elaborate on the Benchmarks, Baselines and Generators currently available in the Routing Arena. We aim to continuously update all three entities and in particular the data generators. All code, data and models to reproduce our experiments and work with the Routing Arena will be available on Github soon. 
\paragraph{Benchmarks.} Given that there already exists some libraries that collect important CVRP benchmarks, the RA currently includes the possibility to evaluate on practically all euclidean-distance benchmarks available on the CVRPLib page. 
Table \ref{tab:benchmarks} describes all euclidean-distance benchmarks available on CVRPLib as well as the concurrent ML Benchmarks. The lower part of the table lists data subsets 
that are currently used for benchmark-testing together with the smaller test sets in \cite{hottung2019neural}. All other benchmark sets are available for individual evaluation except for the duration-constrained CVRP set in \cite{li2005very} and the very large instance set provided by \cite{arnold2019efficiently}, which are left to be included in upcoming versions of the Routing Arena. Notably, the problem sizes in \cite{arnold2019efficiently} are currently not manageable by most ML methods and have only been considered in \cite{li2021learning}, which will also be featured as a baseline in the next version of the RA.
\begin{table}
    \centering
    \resizebox{\columnwidth}{!}{
    \begin{tabular}{l|ccccccl}
    Benchmark & Area & \# Instances & Problem Size & Use in RA & BKS & synthetic & Particularities \\
    \midrule
       Set A \citet{augerat1995computational}   & OR & 27 & 31-79 & indiv. Testing & \blackcheck & synth. & All optimally solved \\
       Set B \citet{augerat1995computational}   & OR & 23 & 30-77 & indiv. Testing & \blackcheck & synth. & All optimally solved \\
       Set E \citet{christofides1969algorithm}  & OR & 13 & 12-100 & indiv. Testing & \blackcheck & synth. & All optimally solved \\
       Set F \citet{fisher1994optimal}          & OR & 3 & 44-134 & indiv. Testing & \blackcheck & synth. & All optimally solved \\
       Set M \citet{christofides1979}           & OR & 5 & 100-199 & indiv. Testing & \blackcheck & synth. & All optimally solved \\
       Set P \citet{augerat1995computational}   & OR & 24 & 15-100 & indiv. Testing & \blackcheck & synth. & All optimally solved \\
       \citet{christofides1979}                 & OR & 14 & 49-198 & indiv. Testing & \blackcheck & synth. & All optimally solved \\
       \citet{rochat1995probabilistic}          & OR & 13 & 75-385 & indiv. Testing & \blackcheck & synth. & All optimally solved \\
       \citet{golden1998impact}                 & OR & 20 & 240-420 & indiv. Testing & \blackcheck & synth. & equiv. solution groups\\
       \citet{li2005very}                       & OR & 12 & 560 - 1200 & to be implemented & \blackcheck & synth & Duration-constr. CVRP\\
       \citet{uchoa2017new}                     & OR & 1000 & 100-1000 & indiv. Testing & \blackcheck & synth. &\\
       \citet{kool2018attention}                & ML & 10000 & 20 - 100 & indiv. Testing & \greencheck & synth. & \citet{nazari2018reinforcement} distrib.\\
       \citet{arnold2019efficiently}            & OR & 10 & 3000-30000 & to be implemented & \blackcheck & real-world & \\
       \citet{hottung2019neural}                & ML & 340 & 100 - 297 & Benchmark-Test & \greencheck & synth. & \citet{uchoa2017new} distrib.\\
       \citet{kool2022deep}                     & ML & 10000 & 100 & indiv. Testing & \greencheck & synth. & \citet{uchoa2017new} distrib.\\
       \citet{queiroga202210}                   & OR & 10000 & 100 & indiv. Testing & \blackcheck & synth. & All optimally solved\\
       \midrule
       \textit{Unif100} & ML & 128 & 100 & Benchmark-Test & \greencheck & synth. & Subset  \citet{kool2018attention}\\
       \textit{XML\_small} & ML & 378 & 100 & Benchmark-Test & \greencheck & synth. & Subset  \citet{queiroga202210}\\
    \end{tabular}}
    \caption{Benchmarks currently included in the Routing Arena. Chronologically from old to new. For each benchmark set, the table lists their use in the RA, as well as the general specifications and particularities.}
    \label{tab:benchmarks}
\end{table}%
As discussed in \cite{uchoa2017new} the instance sets A, B, E, F, M, P and the instances in \cite{christofides1979} and \cite{rochat1995probabilistic} are to a great extent exhausted in terms of their benchmarking capabilities as they are optimally solved and thus too simple to be incorporated for the Benchmark-Test. Furthermore, even though the instances in \cite{golden1998impact} are not all optimally solved, their artificiality and homogeneity make these instances not ideal for benchmarking. 
\paragraph{Generators.} To get the best performance out of ML methods, the methods should be trained on instances that stem from the same distribution as the benchmark instances they are evaluated on. The Routing Arena features configurable data samplers that generate customizable training instances on the fly for Reinforcement Learning based methods. The currently implemented generators include \textit{uniform}-, \textit{gaussian mixture}- and \textit{X}- coordinate distributions as well as \textit{uniform}, \textit{gamma-} and \textit{X}-type demand distributions. 
\paragraph{Baselines.} The baselines currently implemented in the Routing Arena (Table \ref{tab:baselines}) consist of representative, basic, as well as state-of-the-art OR and ML methods. 
Given that we are evaluating based on new protocols and on new benchmarks, we may expect to see performance shifts and therefore do include earlier baselines. 
The methods can roughly be divided into Construction (C) and LS approaches. For the construction approaches, the RA constitutes the option of running a LS after construction in terms of the google OR-Tools (ORT) Large Neighborhood Search with Simulated Annealing (SA). 
Concerning OR methods, there are currently four open-source algorithms included in the RA, two of which are established and well-known in the ML community; the Clark-Wright Savings \cite{savings, rasku2019meta} algorithm and the LKH3 \citet{helsgaun2017extension} heuristic search, and two of which are currently amongst the state-of-the-art heuristic solvers for the CVRP, FILO \cite{accorsi2021fast} and HGS \cite{vidal2012hybrid, vidal2022hybrid}. To get an overview over all individual methods, see appendix \ref{app:baselines}.
\begin{table}[htb]
    \centering
    \resizebox{\columnwidth}{!}{
    \begin{tabular}{l|cccll}
    Baseline & Area & Training Type & Method Type & Method Variants & Venue \\
    \midrule
       Savings-CW \cite{savings}            & OR & -    & C             & +SA           & Operations Research, 1964\\
       LKH \cite{helsgaun2017extension}     & OR & -    & LS            &               & Technical Report, 2017\\
       AM \cite{kool2018attention}          & ML & RL   & C             & +SA           & ICLR 2019\\
       ORT \cite{ortools}                   & OR & -    & LS            & SA, GLS, TS   & Technical Report, 2019\\
       MDAM \cite{xin2021multi}             & ML & RL   & C             & +SA           & AAAI, 2021\\
       NLNS \cite{hottung2019neural}        & ML & RL   & LS            &               & ECAI, 2020\\
       POMO \cite{kwon2020pomo}             & ML & RL   & C             & +SA           & NeurIPS 2020\\
       DACT \cite{ma2021learning}           & ML & RL   & LS            &               & NeurIPS2021\\
       FILO \cite{accorsi2021fast}          & OR & -    & LS            &               & Transportation Science, 2021\\
       NeuroLKH \cite{xin2021neurolkh}      & ML & Supervised    & LS   &               & NeurIPS, 2021\\
       SGBS \cite{choo2022simulation}       & ML & RL   & C             & + EAS         & NeurIPS, 2022\\
       HGS \cite{vidal2022hybrid}           & OR & -    & LS            &               & Computers \& Operations Research, 2022 \\
       NeuroLS \cite{falkner2023learning}   & ML & RL   & LS            &               & ECML PKDD, 2022\\

    \end{tabular}}
    \caption{Description of Baselines included in RA. Chronologically from old to new.}
    \label{tab:baselines}
\end{table}%
\section{Run-Time Normalization}
\label{time_normalization}
A major obstacle for comparability in CO is the measurement of run-time.
While in OR, it is common to evaluate methods on the same infrastructure, usually composed of a single CPU thread, this setting is not useful for the ML field, due to the increased variability of hardware and because the same GPU hardware would not be available to everyone.
To fairly compare run-time efficiency, the RA employs a calibration scheme based on the \emph{PassMark} \footnote{https://www.passmark.com/} hardware rating to normalize time budgets and run-times in the evaluation process.
Similar to the 12th DIMACS challenge \cite{dimacs}, the per instance runtime budget $T_{\scaleto{\text{MAX}}{3.5pt}}$ for a particular machine is normalized to a new budget $\tilde{T}_{\text{max}}$, by multiplying with the ratio of the machines \emph{PassMark} score $s$ and a \textbf{reference machines} \emph{PassMark} score $s_{\text{base}}$.
\begin{equation}
\label{adjusted_timelimit}
         \tilde{T}_{\text{max}} =  T_{\scaleto{\text{MAX}}{3.5pt}} \frac{s_{\text{base}}}{s}
\end{equation}

Analogously, after running the algorithm and retrieving a solution at time $t$, the time is renormalized as if it were run on the reference machine:
\begin{equation}
\label{adjusted_time}
         \tilde{t} = t \frac{s}{s_{\text{base}}}
\end{equation}

Thus, the run-times for evaluation are standardized to a particular reference machine. For methods that are designed to run on a single threaded CPU, we define the mark of $s_{\text{base}}=2000$ as reference machine, similar to the DIMACS challenge. 
Concerning GPU usage, we opt for a standardization scheme incorporating both, CPU and GPU marks. The \emph{PassMark} website issues average G3D and G2D Marks to rate video card performances, we include both with equal weight and linearly combine it with the CPU performance to get the \emph{PassMark} score $s$:
\begin{equation}
\label{GPU_passmark}
         s := \frac{1}{2}(\# \text{CPU} \cdot \text{CPU\_Mark} + \# \text{GPU} \cdot \frac{1}{2}(\text{G3D} + \text{G2D}))
\end{equation}

The reference machine for GPU-usage is set to the combination of a single CPU thread and a single GeForce GTX 1080 machine. Thus, the run-time budget for a single processor and GPU run would be standardized by the value of $s_{\text{base}}=9960$. Moreover, since the \emph{PassMark} website issues different CPU-Marks for single- and multi-thread CPUs, the integrated baselines are monitored for multi-core utilization, such that they are assigned the single-thread CPU-Mark or the multi-thread CPU-Mark accordingly to the utilization in the run. Examples are presented in appendix \ref{app:run_time_stdization}.
We note that this first approach for standardizing run-times in NCO is not yet optimal. Ideally, one would want to have a single $s_{\text{base}}$ for any infrastructure used in the evaluation. However, finding common ground between CPU and GPU performance measurements is not trivial. In the first version of the Routing Arena, we consider separate normalization schemes for GPU- and CPU-based methods and aim to improve on this aspect in future versions.
%
%
\section{Evaluation}
\label{evaluation}
\subsection{Evaluation Protocols}
\paragraph{Fixed Budget.}
The Fixed-budget protocol compares algorithms for an a priori fixed time budget. Hence, for the best obtained solution value $z$ after terminating with a time budget $T_{\scaleto{\text{MAX}}{3.5pt}}$, we compare methods in terms of the relative percentage gap to a BKS with solution value $z_{BKS}$ as follows:
\begin{equation}
\label{fixed_budget_eqn}
         \text{Gap}_{T_{\scaleto{\text{MAX}}{3.5pt}}}(z) = 100 \frac{z - z_{BKS}}{z_{BKS}}
\end{equation}


%
%
\paragraph{Any-time.}
In contrast to the Fixed Budget Evaluation, the "any-time" protocol assesses how well a method performs over its full solution trajectory until the time budget is reached. This highlights the trade-off between run-time and solution quality, and thus returns a more fine-grained performance signal with respect to efficiency. The name relates to the purpose of this problem setting, where practically, one might want to stop the search process at any time and retrieve a fairly decent solution. Many ML-based methods claim to be more "efficient" (\cite{choo2022simulation, xin2021neurolkh, ma2021learning}), however, their evaluation protocol was so far not laid out to validate this. Section \ref{metrics} introduces the metrics to assess any-time performance and thus to \textit{quantifying the aggregated any-time performance} of methods.
\subsection{Metrics}
\label{metrics}
\paragraph{Normalized Primal Integral.}
The Primal Integral (PI) as defined in \cite{berthold2013measuring} evaluates the solution quality development over a method’s optimization process. We briefly introduce the version used in the DIMACS challenge \cite{dimacs}. Given a solution time budget $T_{\scaleto{\text{MAX}}{3.5pt}} \in \R_{\geq 0}$, a sequence of incumbent (i.e. intermediate) solutions with objective costs $z_i$ for $i \in 1,\ldots,n-1$ found at time $t_i \in [0,T]$ where $t_0 = 0$ and $t_n=T$, the average solution quality of a search trajectory in terms of the PI is defined as follows:
%
%
\begin{equation}
\label{PI_dimacs}
         \hspace{0.1 cm}
         PI= 100 \times (\frac{\sum_{i=1}^{n} z_{i-1}\cdot (t_i-t_{i-1}) + z_n \cdot (T_{\scaleto{\text{MAX}}{3.5pt}}-t_n)}{T_{\scaleto{\text{MAX}}{3.5pt}} z_{\text{BKS}}} - 1)
\end{equation}
$PI$ decreases in two cases; (1) a better solution is found at a given time step or (2) a given solution is found at an earlier time step. Thus better PI scores are expected for methods that deliver competitive solutions fast. However, we see two points for improvement; (a) We are often interested in the relative performance increase over a basic any-time search heuristic that often in ML functions as a naive baseline demonstrating the learning behavior. (b) The weight on the final solution, ($T_{\scaleto{\text{MAX}}{3.5pt}} - t_n$), encourages methods to find the last solution at time points close to $T_{\scaleto{\text{MAX}}{3.5pt}}$. This attribute is not necessary for any-time performance and would disadvantage construction heuristics that find only a single solution early on. 

%

%
\paragraph{Weighted Relative Average Performance.}
We propose to measure the any-time performance with the Weighted Relative Average Performance (WRAP), which reflects the same principles as the PI, but refines the quality measurement in the two aspects outlined above. It is based on the relative performance improvement ($RPI$) over a \textbf{base solver}. In the current version of the Routing Arena, the Clark and Wright Savings Algorithm together with a SA-guided ORT search is chosen as base solver, as both the Savings and the SA procedure stand for being simple and effective in terms of solving routing problems.
At a given time $t$ of the search, the $RPI$ for a method with objective cost $z_t$, a corresponding BKS $z_{\text{BKS}}$ and a base solution with cost $z_{t}^{\text{base}}$ is defined as follows:

\begin{equation}
\begin{aligned}
\label{RPI}
RPI(t) := 
     \begin{cases}
         1, & \text{if} \quad t=0 \text{ (no incumbent solution so far)}\\
    \frac{\min\{z_{t}, z_{t}^{\text{base}}\}\ - \ z_{\text{BKS}}}{z_{t}^{\text{base}}\ - \ z_{\text{BKS}}}, & \text{for incumbent solution costs at time $t$}\\
     \end{cases}   
\end{aligned}
\end{equation}

Given equation \ref{RPI}, the $RPI$ remains at a value of 1, as long as the incumbent solution cost $z_t$ is worse (larger) than the base solution cost and only decreases for incumbents with cost $z_t < z_{t}^{\text{base}}$. Furthermore, the run-times for the benchmarked solver’s solutions $z_{t}$ and the base solver’s solutions $z_{t}^{\text{base}}$ are synchronized in a pre-processing step of the $RPI$ calculation, where for each time $t$, at which a solver found a new solution, while the base solver did not, we use the base solver’s best solution up until time point $t$ in order to calculate the $RPI$.

The WRAP $\in [0, 1]$ for a method with incumbent solutions $x_i, i \in 1,\ldots, n$ found at time $t_i$ until the time limit $T_{\scaleto{\text{MAX}}{3.5pt}}$ is reached and $t_0 = 0$, is defined as:
\begin{equation}
\begin{aligned}
     WRAP :=  \frac{1}{T_{\scaleto{\text{MAX}}{3.5pt}}} \sum_{i=1}^{n} RPI(t_{i-1}) (t_{i} \ - \ t_{i-1})
\end{aligned}
\end{equation}
The attributes of WRAP can be summarized as follows; (1) It incorporates the time until initial solution explicitly, since the $RPI$ is equal to one for the initial solution provided at time $t_0$. (2) The comparison to a simple and popular heuristic, Savings+SA, measures the typical performance gain over a simple baseline required for a consistent evaluation of ML and OR methods. (3) The WRAP metric does not systematically favor methods that find solutions close to the termination criteria. (4) Because WRAP compares to a base solution and a BKS, it aggregates well over heterogeneous instance sets.

\section{Experiments and Evaluation}
\label{experiments}

In this section, we engage in representative experiments that incorporate the uniformly distributed \textit{Unif100} instances from \cite{kool2018attention}, new instances from \cite{queiroga202210} (\textit{XML\_small}), and the \textit{X}-type distributed \textit{XE} instances provided by \cite{hottung2019neural}. 
Each of the \textit{XE} sets consists of a specific distribution in the \textit{X-set} (see section \ref{related_work}).
Most ML methods presented in Table \ref{table:gap1}, \ref{table:pi1} and \ref{table:wrap1} are retrained for the \textit{X}-type distribution, while the originally trained model checkpoints are provided for the uniform data-distribution.

  \begin{minipage}[b]{0.55\textwidth}  
    \centering
    \resizebox{\columnwidth}{!}{
    \renewcommand{\arraystretch}{1} 
    \setlength\tabcolsep{2pt}
  \begin{tabular}{l|lllll|llll}  
    \toprule
    \textbf{Gap (\%)} & \multicolumn{5}{c|}{ML} & \multicolumn{4}{c}{OR}  \\
    Dataset & Pomo   & Neu.  & Neu.  & SGBS  & NLS   & CW    & LKH   & FILO  & HGS  \\
            & +SA    & LNS   & LKH         & +EAS  &   & +SA    &   &  &  \\
    \midrule
    Unif100                  	&	1.58	&	188	    &	1.09	&	\textit{\underline{0.22}}	&	2.41	&	4.09	&	1.14	&	0.955	&	 \textbf{0.0058} \\
    XML                      	&	3.53	&	164	    &	\textit{0.23}	&	12.6	&	3.29	&	5.48	&	0.18	&	\underline{0.061}	&	 \textbf{0.0258} \\
    XE\textsubscript{1}      	&	1.79	&	1.65	&	\textit{0.54}	&	1.23	&	1.07	&	3.48	&	0.46	&	\underline{0.160}	&	 \textbf{0.0007} \\
    XE\textsubscript{2}      	&	2.42	&	3.01	&	\textit{1.07}	&	2.11	&	1.43	&	4.91	&	0.96	&	\underline{0.314}	&	 \textbf{0.0013} \\
    XE\textsubscript{3}      	&	1.63	&	2.99	&	\textit{0.50}	&	0.65	&	3.28	&	2.64	&	0.42	&	\underline{0.087}	&	 \textbf{0.0045} \\
    XE\textsubscript{4}      	&	2.56	&	4.04	&	\textit{0.26}	&	1.20	&	2.86	&	6.88	&	0.29	&	\underline{0.065}	&	 \textbf{0.0016} \\
    XE\textsubscript{5}      	&	1.60	&	3.60	&	\textit{0.04}	&	1.53	&	1.10	&	2.02	&	0.05	&	\underline{0.004}	&	 \textbf{0.0034} \\
    XE\textsubscript{6}      	&	3.91	&	7.85	&	\textit{0.49}	&	2.05	&	3.55	&	6.19	&	0.47	&	\underline{0.149}	&	 \textbf{0.0051} \\
    XE\textsubscript{7}      	&	2.98	&	6.92	&	\textit{0.75}	&	2.27	&	1.72	&	4.44	&	0.75	&	\underline{0.176}	&	 \textbf{0.0002} \\
    XE\textsubscript{8}      	&	4.20	&	6.29	&	\textit{0.51}	&	2.74	&	3.02	&	5.76	&	0.39	&	\underline{0.076}	&	 \textbf{0.0005} \\
    XE\textsubscript{9}      	&	4.39	&	9.29	&	\textit{0.76}	&	3.62	&	3.34	&	5.84	&	0.66	&	\underline{0.055}	&	 \textbf{0.0092} \\
    XE\textsubscript{10}     	&	0.50	&	0.20	&	\textit{0.01}	&	1.17	&	47.0	&	0.41	&	0.01	&	\underline{0.002}	&	 \textbf{0.0005} \\
    XE\textsubscript{11}     	&	4.16	&	4.44	&	\textit{0.30}	&	5.01	&	8.79	&	5.97	&	0.43	&	\underline{0.051}	&	 \textbf{0.0005} \\
    XE\textsubscript{12}     	&	2.87	&	4.53	&	\textit{0.79}	&	3.15	&	4.18	&	2.20	&	0.79	&	\underline{0.130}	&	 \textbf{0.0094} \\
    XE\textsubscript{13}     	&	6.06	&	6.73	&	\textit{1.15}	&	7.57	&	2.61	&	4.53	&	0.75	&	\underline{0.179}	&	 \textbf{0.0000} \\
    XE\textsubscript{14}     	&	3.85	&	5.42	&	\textit{0.42}	&	4.47	&	11.0	&	2.58	&	0.19	&	\underline{0.015}	&	 \textbf{0.0053} \\
    XE\textsubscript{15}     	&	5.78	&	7.28	&	\textit{1.69}	&	6.88	&	3.84	&	4.29	&	1.15	&	\underline{0.402}	&	 \textbf{0.0001} \\
    XE\textsubscript{16}     	&	6.96	&	4.36	&	\textit{1.17}	&	7.94	&	2.13	&	3.41	&	0.94	&	\underline{0.212}	&	 \textbf{0.0008} \\
    XE\textsubscript{17}     	&	7.93	&	8.86	&	\textit{1.03}	&	6.03	&	3.81	&	4.25	&	1.04	&	\underline{0.239}	&	 \textbf{0.0019} \\

    \bottomrule
    AVG                      & 3.62 &	23.3 &	\textit{0.67} & 3.81 & 5.81 &	4.18 &	0.58 &	\underline{0.175} &	\textbf{0.0040} \\
    \bottomrule
  \end{tabular} 
  }
    \captionsetup{type=table, width=.9\linewidth}
    \captionof{table}{Relative Gap (in \%). Results are averaged over three runs for a time budget of $T_{\scaleto{\text{MAX}}{3.5pt}} = 2.4N$ (s). Standard Deviations are available in the appendix \ref{app:furtherResults}. \textbf{Best}, \underline{2nd best} and \textit{best ML} approach.}
    \label{table:gap1}
    \end{minipage}
    \begin{minipage}[b]{0.43\textwidth}
    \centering
    \captionsetup{type=figure, width=.9\linewidth}
    \includegraphics[trim= {0cm 0cm 0cm 0cm}]{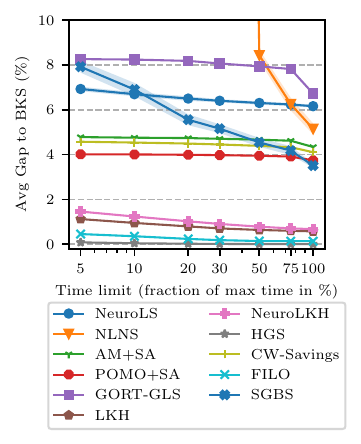}
    \captionof{figure}{Relative Gap to the BKS vs Time Limit averaged over all XE sets. The shaded area represents $\pm$2 standard deviations from 3 runs.}
    \label{pareto_XE}
  \end{minipage}
\subsection{Fixed-Budget Performance}
For fixed budget evaluation, we present the gap to the BKS in table \ref{table:gap1} for the max time budget and in figure \ref{pareto_XE} over different time budgets. The unnormalized time budgets are determined similarly to \cite{vidal2022hybrid} by instance size ($T_{\scaleto{\text{MAX}}{3.5pt}} = 2.4N$), except for \textit{Unif100}, where we consider 120 seconds to be sufficient.

Table \ref{table:gap1} shows that for the max. time budget most ML-based methods show worse average performance compared to state-of-the-art OR heuristics and only POMO+SA, NeuroLKH and SGBS+EAS perform on average better than the designated base solver Savings+SA. Furthermore, NeuroLKH does not outperform its base heuristic LKH on average, but does marginally outperform it on separate instance sets, for which it was trained (\textit{Unif100}, and some of the \textit{XE} sets), thus is able to learn from the training distributions in order to improve on LKH.
NLNS shows particularly bad results for \textit{Unif100} and \textit{XML} even though it is trained on uniform data. This is due to the NLNS "single-instance" algorithmic setting described in appendix \ref{app:baselines} and the fact that NLNS is very expensive in terms of computational resources, leading to a reduced normalized runtime budget. 
Figure \ref{pareto_XE} shows that there are effectively two performance categories of methods. Ones that show percentage gaps below 2\% for even a time limit of 5\% of the max time and those that barely cross the 4\% gap to the BKS when the time limit is maximal. 
The figure also shows that many methods only marginally improve when increasing the time budget which shows that many OR baselines such as LKH, that are often attributed run-times magnitudes larger than those of ML methods \cite{kool2018attention, ma2021learning, kwon2020pomo} find superior solutions early on in the search. 
Especially the implementation of HGS (HGS-CVRP), developped by \citet{vidal2012hybrid}, demonstrates convincing results with regards to computational efficiency, which is due to the implementation's highly optimized code in C++ and adapted and trimmed down data structures.
There are two exceptions to this constant performance behaviour; NLNS, which only shows significant performance gains for time limits that are greater than fifty percent of the maximum time budget and SGBS-EAS, for which we see inferior performance for small time limits but can account for the forth best fixed budget performance concerning the maximal time budget.
\subsection{Any-time Performance}
\begin{table}
    \caption{Anytime metrics for a selection of methods and datasets. The results are averaged over three runs for a time budget in seconds of $T_{\scaleto{\text{MAX}}{3.5pt}} = 2.4N$. Standard deviations are usually small and thus omitted from this table. They are available in the appendix \ref{app:furtherResults}. \textbf{Best}, \underline{2nd best} and \textit{best ML} approach.}
    \label{table:anytime}
    \begin{subtable}{.48\linewidth}
      \centering
        \caption{Primal Integral (PI)}
        \label{table:pi1}
        \setlength\tabcolsep{2pt}
    \resizebox{\columnwidth}{!}{
  \begin{tabular}{l|lllll|llll}  
    \toprule
    \textbf{PI} & \multicolumn{5}{c|}{ML} & \multicolumn{4}{c}{OR}  \\
    Dataset & Pomo   & Neu.  & Neu.  & SGBS  & NLS   & CW    & LKH   & FILO  & HGS  \\
            & +SA    & LNS   & LKH         & +EAS  &   & +SA    &   &  &  \\
    \midrule
    Unif100                 & 1.62   & 10   & 1.37 & \textit{\underline{0.57}}	&	2.87	&	4.14	&	1.36	&	1.04 & \textbf{0.012} \\
    XML                     & 2.94   & 10   & \textit{0.37} & 9.04	&	2.99	&	4.55	&	0.27	&	\underline{0.11} & \textbf{0.030}  \\
    XE\textsubscript{1}     & 1.84   & 5.17 & \textit{0.82} & 2.09	&	1.31	&	3.50	&	0.63	&	\underline{0.22} & \textbf{0.007} \\
    XE\textsubscript{2}     & 2.48   & 6.04 & \textit{1.39} & 3.08	&	1.75	&	5.32	&	1.20	&	\underline{0.43} & \textbf{0.011} \\
    XE\textsubscript{3}     & 1.69   & 6.16 & \textit{0.79} & 1.01	&	3.50	&	2.76	&	0.66	&	\underline{0.13} & \textbf{0.027} \\
    XE\textsubscript{4}     & 2.61   & 6.96 & \textit{0.37} & 2.09	&	3.19	&	7.01	&	0.39	&	\underline{0.09} & \textbf{0.012} \\
    XE\textsubscript{5}     & 1.69   & 7.53 & \textit{0.10} & 2.82	&	1.22	&	2.22	&	0.08	&	\underline{0.02} & \textbf{0.006} \\
    XE\textsubscript{6}     & 4.01   & 9.77 & \textit{0.74} & 3.53	&	3.91	&	6.37	&	0.61	&	\underline{0.20} & \textbf{0.021} \\
    XE\textsubscript{7}     & 3.10   & 9.07 & \textit{1.01} & 3.35	&	1.95	&	4.59	&	0.93	&	\underline{0.28} & \textbf{0.017} \\
    XE\textsubscript{8}     & 4.28   & 9.14 & \textit{0.75} & 3.71	&	3.30	&	5.94	&	0.51	&	\underline{0.15} & \textbf{0.009} \\
    XE\textsubscript{9}     & 4.40   & 9.73 & \textit{0.96} & 4.77	&	3.65	&	5.94	&	0.76	&	\underline{0.15} & \textbf{0.025} \\
    XE\textsubscript{10}    & 0.53   & 2.92 & \textit{0.06} & 1.86	&	10.0	&	0.43	&	0.05	&	\underline{0.01} & \textbf{0.003}\\
    XE\textsubscript{11}    & 4.23   & 7.21 & \textit{0.48} & 7.05	&	8.99	&	6.15	&	0.58	&	\underline{0.11} & \textbf{0.027} \\
    XE\textsubscript{12}    & 2.93   & 8.07 & \textit{1.03} & 4.27	&	4.40	&	2.23	&	1.02	&	\underline{0.21} & \textbf{0.088} \\
    XE\textsubscript{13}    & 6.18   & 9.18 & \textit{1.62} & 8.81	&	2.97	&	4.58	&	0.98	&	\underline{0.29} & \textbf{0.029} \\
    XE\textsubscript{14}    & 3.93   & 7.33 & \textit{0.67} & 5.78	&	10.0	&	2.62	&	0.26	&	\underline{0.07} & \textbf{0.019} \\
    XE\textsubscript{15}    & 5.82   & 8.93 & \textit{2.11} & 7.76	&	4.12	&	4.34	&	1.51	&	\underline{0.56} & \textbf{0.055} \\
    XE\textsubscript{16}    & 7.07   & 6.80 & \textit{1.58} & 9.07	&	2.36	&	3.45	&	1.21	&	\underline{0.33} & \textbf{0.051} \\
    XE\textsubscript{17}    & 8.02   & 9.86 & \textit{1.23} & 7.73	&	4.07	&	4.29	&	1.22	&	\underline{0.38} & \textbf{0.058} \\
    \bottomrule
    AVG	                    & 3.65   & 7.89	& \textit{0.92} & 4.65   & 4.03      & 4.23      & 0.75      & \underline{0.25}   & \textbf{0.027} \\
    \bottomrule
  \end{tabular}
  }
    \end{subtable}%
    \begin{subtable}{.48\linewidth}
      \centering
        \caption{Weighted Relative Average Performance (WRAP)}
        \label{table:wrap1}
        \setlength\tabcolsep{2pt}
    \resizebox{\columnwidth}{!}{
  \begin{tabular}{l|lllll|llll}  
    \toprule
    \textbf{WRAP} & \multicolumn{5}{c|}{ML} & \multicolumn{4}{c}{OR}  \\
    Dataset & Pomo   & Neu.  & Neu.  & SGBS  & NLS   & CW    & LKH   & FILO  & HGS  \\
            & +SA    & LNS   & LKH         & +EAS  &   & +SA    &   &  &  \\
    \midrule
         Unif100                 	&	0.40	&	1	    &	0.35	&	\underline{\textit{0.14}}	&	0.70	&	0.94	&	0.35	& 0.254 & \textbf{0.003} \\
         XML                     	&	0.56	&	1	    &	\underline{\textit{0.10}}	&	1	    &	0.58	&	0.92	&	0.13	& \textbf{0.027} & 0.291 \\
         XE\textsubscript{1}     	&	0.55	&	0.84	&	\textit{0.25}	&	0.60	&	0.40	&	0.98	&	0.18	& \underline{0.120} & \textbf{0.008}  \\
         XE\textsubscript{2}     	&	0.47	&	0.83	&	\textit{0.27}	&	0.58	&	0.35	&	0.94	&	0.23	& \underline{0.091} & \textbf{0.002} \\
         XE\textsubscript{3}     	&	0.62	&	0.99	&	\textit{0.29}	&	0.37	&	0.98	&	0.96	&	0.23	& \underline{0.056} & \textbf{0.009} \\
         XE\textsubscript{4}     	&	0.37	&	0.85	&	\textit{0.05}	&	0.30	&	0.45	&	0.98	&	0.07	& \underline{0.012} & \textbf{0.002} \\
         XE\textsubscript{5}     	&	0.73	&	1	    &	\textit{0.06}	&	0.91	&	0.56	&	0.98	&	0.03	& \underline{0.009} & \textbf{0.003} \\
         XE\textsubscript{6}     	&	0.63	&	1	    &	\textit{0.12}	&	0.55	&	0.61	&	0.99	&	0.10	& \underline{0.033} & \textbf{0.003} \\
         XE\textsubscript{7}     	&	0.67	&	1	    &	\textit{0.21}	&	0.69	&	0.43	&	0.99	&	0.20	& \underline{0.069} & \textbf{0.004} \\
         XE\textsubscript{8}     	&	0.72	&	0.99	&	\textit{0.13}	&	0.62	&	0.56	&	1	    &	0.09	& \underline{0.027} & \textbf{0.001} \\
         XE\textsubscript{9}     	&	0.73	&	1	    &	\textit{0.16}	&	0.78	&	0.63	&	0.99	&	0.12	& \underline{0.028} & \textbf{0.005} \\
         XE\textsubscript{10}    	&	0.92	&	0.84	&	\textit{0.07}	&	1	    &	1	    &	0.96	&	0.05	& \underline{0.010} & \textbf{0.006} \\
         XE\textsubscript{11}    	&	0.68	&	0.92	&	\textit{0.08}	&	0.94	&	1	    &	0.97	&	0.09	& \underline{0.018} & \textbf{0.004} \\
         XE\textsubscript{12}    	&	0.98	&	1	    &	\textit{0.46}	&	1	    &	1	    &	0.97	&	0.45	& \underline{0.130} & \textbf{0.040} \\
         XE\textsubscript{13}    	&	0.98	&	1	    &	\textit{0.35}	&	1	    &	0.65	&	0.96	&	0.21	& \underline{0.073} & \textbf{0.006} \\
         XE\textsubscript{14}    	&	0.95	&	1	    &	\textit{0.23}	&	1	    &	1	    &	0.95	&	0.08	& \underline{0.032} & \textbf{0.008} \\
         XE\textsubscript{15}    	&	0.95	&	1	    &	\textit{0.49}	&	1	    &	0.87	&	0.97	&	0.35	& \underline{0.180} & \textbf{0.012} \\
         XE\textsubscript{16}    	&	1	    &	1	    &	\textit{0.45}	&	1	    &	0.70	&	0.99	&	0.35	& \underline{0.166} & \textbf{0.015} \\
         XE\textsubscript{17}    	&	1	    &	1	    &	\textit{0.30}	&	1	    &	0.89	&	0.98	&	0.31	& \underline{0.135} & \textbf{0.015}  \\
    \bottomrule
    AVG                     & 0.73    & 0.96    & \textit{0.23} &	0.76   & 0.70   & 0.97 & 0.19 & \underline{0.077} & \textbf{0.023}  \\
    \bottomrule
  \end{tabular}
  }
    \end{subtable} 
\end{table}%
For the any-time performance, we engage in the same experimental setup regarding method selection and present the two metrics, PI and WRAP, in tables \ref{table:pi1} and \ref{table:wrap1} to demonstrate the run-time efficiency of methods. Note that the PI metric has values ranging from zero to ten, while WRAP lies between zero and one. For both, smaller is better.

Similarly to the fixed-budget evaluation, we see a clear winner in both the PI and WRAP scores in the highly efficient HGS-CVRP \cite{vidal2022hybrid}, again followed by FILO \cite{accorsi2021fast} and LKH \cite{helsgaun2017extension}. Concerning ML-based methods, the ranking of baselines in the average any-time evaluation is consistent with the fixed budget evaluation, but the metric values for the individual sets are more diverse especially concerning SGBS+EAS, POMO+SA and NLS. Furthermore, as the instances sizes increase from XE\textsubscript{1} to XE\textsubscript{17}, we see a more pronounced decrease in any-time performance amongst neural solvers, especially in the WRAP metric.
Compared to PI, WRAP is stricter in the sense that it attributes a score of 1.0 more often than a PI of 10, since in WRAP evaluation, the neural solvers are competing with a base solver that yields quality results for some instance sets. 
From the results above, we see that the neural solvers that deliver better any-time performance on average and in a more stable fashion are hybrid ML-OR algorithms, such as NeuroLKH and POMO+SA, while pure neural search approaches are not competitive, except NLS, which yields the second best WRAP score amongst ML-based methods on average. Looking at the sets individually, SGBS-EAS reaches the second best any-time performance after HGS for uniformly distributed data, which is presented also more clearly in Fig. \ref{fig:trajectories_unif100}.
\section{Conclusion and Future Work}
\label{conclusion}
The last few years of NCO have shown that learning to optimize routing problems holds great potential for circumscribing the need for hand-crafted, expert-knowledge intensive algorithm development. However, this work pinpoints essential shortcomings in the way this recent progress has been framed and evaluated and proposes the \textit{Routing Arena} as first step towards a consistent two-folded evaluation protocol (any-time and fixed budget) and development tool to promote research in both, the OR- and ML research field. To this end, we see that future work will be required in extending the capabilities to further routing problem variants, problem sizes and furthermore, in exploring new, even more challenging data distributions to train and evaluate NCO methods.

\newpage
\bibliography{neurips_data_2023}





\newpage


\appendix

\lstset{frameround=fttt,language=python,numbers=left,breaklines=true}


\section{Benchmark Suite Design \& Structure}
\label{design}

The choice of design for the Benchmark Suite is influenced by the repository being able to cater to both, researcher's needs for efficient testing and benchmarking as well as the broader audiences' interest in the NCO topic. Therefore, we establish two main \textit{desiderata}: \textbf{Simplisticity} in terms of the structure being easy to understand and clearly documented, featuring default settings that don't need a lot of prior and expert knowledge to be executed and \textbf{Sophisticated Configuration Management} that leaves room for optionality and technical adjustments in terms of baseline control and detailed experiment setups. The following subsections document (1) the structure of the repository and formats of important entities, (2) the benchmark run configuration management, (3) a guide of how baselines can be integrated to perform evaluation tests, (4) quick-start directions on how to reproduce experiments and lastly (5) a closer look at and description of the algorithms and models already implemented in the Routing Arena.

\subsection{Benchmark Structural Design}

The Routing Arena repository covers three main entities (directories): \textbf{data}, \textbf{models} and \textbf{outputs}, which will be elaborated on in this section.

\paragraph{Data.} The data directory most importantly covers the benchmarks for the problem at hand that are ordered i.t.o. data-distributions. Each of the benchmarks (see Table \ref{tab:benchmarks}) is stored in its original form, so that for the ML test datasets are stored in \texttt{.pt} or \texttt{.pkl} format, whereas most of the OR-domain datasets are stored as text-like \texttt{.vrp} files. The data loading functionality of the RA brings the different formats of the test (or on-the-fly generated) data into a single \texttt{namedtuple()} input format called \textit{CVRPInstance} (Listing \ref{lst:CVRPInstance}) incorporating necessary information on the \textit{problem}, but also \textit{configuration choices}, such as a particular per-instance time limit in seconds, \textit{distribution characteristics} and the \textit{BKS} if available.
Similarly, the complete solution information of an evaluation run is formatted in an \textit{RPSolution} format (Listing \ref{lst:RPSolution}), where the final \textit{solution} as well as the "\textit{running\_sols}" (e.g. incumbent solutions) and the metric scores that summarize the run's performance are stored.
Examples of the input and output structure are listed below:
\footnotesize
\begin{lstlisting}[caption={Routing Arena Instance Format}, label={lst:CVRPInstance},numbers=none, language=python]
CVRPInstance( coords=ndarray_[101, 2],
              node_features=ndarray_[101, 5],  
              graph_size=101,  
              vehicle_capacity=1.0,  
              original_capacity=206,  
              max_num_vehicles=None,  
              depot_idx=[0],  
              constraint_idx=[-1],  
              time_limit=240,  
              BKS=21601.80798,  
              instance_id=18,  
              coords_dist=RC, 
              depot_type=R, 
              demands_dist=1-100,  
              original_locations=ndarray_[101, 2],  
              type=XE_1)
\end{lstlisting}
\begin{lstlisting}[caption={Routing Arena Solution Format}, label={lst:RPSolution},numbers=none, language=Python]
RPSolution( solution=[[0, 40, 16, 21, 39, 0],...,[0, 56, 59, 41, 0]], 
            cost=34189.378, 
            pi_score=2.691, 
            wrap_score=0.690, 
            num_vehicles=24, 
            run_time=240, 
            problem='CVRP', 
            instance=CVRPInstance(coords=ndarray_[101, 2],...),
            running_costs=[35199.873, ..., 34189.378], 
            running_times=[0.124,..., 105.746], 
            running_sols=[[[...]]])
\end{lstlisting}
\normalsize

For each metric evaluation run, the benchmark set's Base Solutions and BKS are retrieved from \texttt{pickle} files, which are \texttt{Dictionaries}, such that the BKS can be updated straightforwardly and efficiently during an evaluation run. An example of the BKS format is illustrated below in \ref{lst:bks}. Furthermore, we plan to store the BKS information also in respective \texttt{.vrp} files that will be visible at the Routing Arena repository and function as a more readable and tangible source to document the \textit{objective cost}, \textit{solution} and if available the according winning \textit{algorithm} and whether the solution is optimal or not.
\footnotesize
\begin{lstlisting}[caption={Example of the BKS File Format for Instance 18 of set XE\_1},label={lst:bks},numbers=none]
bks_XE1['18'] = ( 21601.80798,
                  [[9, 47, 4, 7],[90, 59, 74, 97],...,[31, 63, 46]],
                  'HGS',
                  'not_opt')
\end{lstlisting}
\normalsize

Note that, currently the RA does not yet assess the optimality of a Best Known Solution, and by default marks the solution as not optimal, the concurrent information of a BKS being optimal is retrieved from CVRPLib. Concerning future use, we will maintain the \texttt{BKS.vrp} files (also in accordance with updates on CVRPLib) and allow the community to contribute new BKS entries via pull requests.

\paragraph{Models.} The baseline codes used in the Routing Arena are slightly modified and/or embedded versions of the originally published source code (\cite{accorsi2021fast, vidal2022hybrid, hottung2019neural, kool2018attention, kool2022deep, kwon2020pomo, xin2021neurolkh, xin2021multi, falkner2023learning, choo2022simulation}) or respective pip library implementations (\cite{ortools,savings}). If changes were conducted in the original code, this is, to the best of our knowledge, noted in the form of comments directly in the code. However, the main source code has mostly not been altered, instead the methods are wrapped and the options and arguments channeled through the the open-source Python framework for hierarchical configuration, Hydra\footnote{https://hydra.cc/docs/configure\_hydra/intro/}.
The baseline directories consist in essence of four entities (two directories and two python files); (1) A \textit{Config} directory, which contains the run-specifications, model-arguments and environment options for a run and will be elaborated on in the next section. (2) A \textit{model} directory consisting of the source code of the particular method termed \texttt{model}. (3) A \texttt{model.py} file, which works as a wrapping element in the Routing Arena, binding the source code and RA functionalities (see subsection \ref{app:baseline_integr}) and (4) a \texttt{runner.py} file that sets up and executes the configured run. This structure is maintained throughout the baselines and is logical to follow through, which naturally simplifies the consolidation of a unified evaluation protocol.

\paragraph{Outputs.} The outputs directory consist of summaries and logs of evaluation and training runs. By default, run information will be grouped in the environment space, such that, for example, all evaluation (val) runs on CVRP-100 uniformly-distributed data for a particular model will be collected in the same directory: \texttt{outputs/cvrp\_100\_uniform/val/model} and separated by date-time directories. If not specified otherwise in the \textit{Config}, the cost trajectory of the first instance is plotted and saved in the directory together with the logs and list of \texttt{RPSolutions} of a run. Furthermore, to make it more convenient to analysed evaluation results in hindsight, the Routing Arena features the option to store result summaries in the \textit{saved\_results} directory that can seamlessly be adopted by the analyser functions in \texttt{analyse.py}.

\subsection{Run Option Management}
\label{app:config_folders}

All arguments regarding the model, algorithm or policy, specifications concerning the test- or train data (environment) and evaluation (run) or train options are handled by the configuration management tool Hydra, which operates on the basis of directories that consist of \texttt{.yaml} files that comprise the run arguments. These directories and files are collected in the method's \textit{Config} folder and can be overwritten via the command line.
In particular, each aspect of a train or test run is configurable as follows; 
\begin{itemize}
    \item \textit{The environment}, specified in the \textit{Config}'s \texttt{env} directory, captures the configuration concerning the problem that is generated for training or loaded for evaluation. For example, the \texttt{cvrp100\_unf.yaml}, contains arguments concerning the coordinate distribution (\texttt{coords\_dist:uniform}) and graph size (\texttt{graph\_size:100}), the path to the default evaluation dataset and to the model checkpoint to use for this environment, as well as potential normalization schemes to be applied on the data for the model to work in this environment. Furthermore, one can also specify the sampling and generator arguments for generating data on the fly, such as the number of samples, number of vehicles and the particularities about the demand distribution in the \texttt{env\_kwargs}. We have pre-configured a couple of the important environments as defaults in the Routing Arena, notably those where metric-evaluations can be performed.
    \item \textit{The policy}, specified in the \textit{Config}'s \texttt{policy} (or \texttt{model}) directory,  is the place for the baseline-specific options and arguments and potentially the reinforcement learning policy arguments. One can choose a single \texttt{.yaml} file and detail out different policy- or model- specifications in this single file or create seperate policy-configuration files for each of the versions and parts of the method. 
    \item \textit{The meta configuration} defines which type of run is to be performed, such as validation, training or debug run. Concerning the evaluation functionality of the Routing Arena, the \texttt{run.yaml} configuration file is the most important, since it configures the test parameters for the run. These test parameters involve amongst others, the time limit defaulted to 10 seconds, options how to save the results of the run, what type of evaluation to perform (whether \texttt{pi} and/or \texttt{wrap} scores should be computed) and whether GPU capabilities should be used.
\end{itemize}
These configuration files are stored for each model in the \texttt{models/model/config} directory together with a config.yaml file that summarizing the configuration choices for the run entities and ships it to the \texttt{runner.py} file.

\subsection{Integrating Baselines}
\label{app:baseline_integr}

The integration of baseline methods follows, for most parts, a systematic setup. There are discrepancies between different methods, but as long as the source of input to the model and the model's output can be easily modified, the integration should be able to work smoothly. That said, some existing RA baselines did not come with the ability to provide a series of incumbent solutions (or at least incumbent costs) at the end of the run (for example NLNS \cite{hottung2019neural} and DACT \cite{xin2021multi}). However this information is necessary to perform the \emph{PI} and \emph{WRAP} evaluations. Hence, for some existing methods we changed the source code in this regards, to be able to include the methods in our analysis. Nevertheless, since most algorithms (ML or OR-based) handle the incumbent solution internally, this adjustment should also be fairly simple to make.
The steps to integrate a baseline with the source-code directory termed \texttt{model} (here the \texttt{model} directory should be the root directory from which the evaluation and training functionality of the \texttt{model} is executed) consists of the following steps:
\begin{enumerate}
    \item Create a directory in the \texttt{models} directory: \texttt{models/model}
    \item Copy the original source code folder into \texttt{models/model}: \vspace{0.2cm} \\
        \phantom{aaa} \texttt{models/model/model} contains now the original source code
    \item Creating a \texttt{model.py} file with the following four functions:
    \begin{enumerate}[label=(\alph*)]
        \item A function that \textit{transforms} \texttt{CVRPInstances} \ref{lst:CVRPInstance} to the \textit{data} format used by \texttt{model}.
        \vspace{0.2cm} \\
         \phantom{aaa} For ML-based methods this data format comes mostly down to the format initiated \phantom{aaa} in \citet{kool2018attention} and can thus be taken from other baseline implementations.
        \item A function that \textit{transforms} the \texttt{model} \textit{solutions} to the \texttt{RPSolution} format \ref{lst:RPSolution}
         \vspace{0.2cm} \\
         \hspace*{0.55cm}%
         \begin{minipage}{.8\textwidth}
         As mentioned earlier, the provided information needs to have a list of incumbent solutions or objective cost and their respective run-times to be eligible for full evaluation. At least a final solution and the respective run-time will be needed to perform any evaluation. 
         \end{minipage}%
        \item A function that calls the model's internal \textit{evaluation} function and processes the transformations (a) and (b).
         \vspace{0.2cm} \\
         \hspace*{0.55cm}%
         \begin{minipage}{.8\textwidth}
         This function will be called by the \texttt{runner.py} once the dataset class (and potentially the model environment) is set up. Note that this is also where the \texttt{model}'s evaluation arguments and options are passed, such as decoding options for a construction method or number of iterations for a search.
         \end{minipage}%
         \item A function that calls the model's internal \textit{training} procedure \textit{if available}.
         \vspace{0.2cm} \\
         \hspace*{0.55cm}%
         \begin{minipage}{.8\textwidth}
         This function can also be called by the \texttt{runner.py} for ML-based methods. Besides passing in training arguments, which are for most parts described in the \texttt{model/config/meta/train.yaml} file, it is necessary to ensure that the transformation function in (a) is passed to the dataset class in the \texttt{runner.py} file, to generate train-data on the fly.
         \end{minipage}%
    \end{enumerate}
    \item Creating a \texttt{runner.py} file to be located in the \texttt{models/model}. We aim to provide two or three template \texttt{runner.py} files  in \texttt{models/template\_model}, which consist of pre-configured code files for (i) construction-type models (with the option of adding a ORT \cite{ortools} LNS on top), (ii) for neural search models and potentially (iii) OR-based methods.
    \item Creating a \texttt{run\_model.py} file located in the root directory. Also this file will be pre-configured and only needs changes with respect to the \textit{Config} directory path and \texttt{runner.py} import. This file will be called to execute a run from the command line.
\end{enumerate}

\subsection{Reproducing Results}
\label{reproduce_results}
The software requirements (python libraries) will be provided as conda environment file (\texttt{environment.yml}) through the Routing Arena repository. Thus a first pre-requisite to reproduce the results will be a python installation through the anaconda distribution \footnote{https://anaconda.org/anaconda/python}. Secondly, it is recommended to use a machine with GPU capabilities regarding the ML-based methods. Even though most of the models also run on CPU, the evaluation run-times will be considerably increased when running on a CPU only machine. Moreover, to reproduce the results of the state-of-the-art OR-baselines (HGS and FILO) we refer users to go through the two- or three-step installation guide noted in the \texttt{Readme.md} documentation of the baseline.\\ 

Given the provided configuration files discussed in \ref{app:config_folders} and the command line overwrite functionality in Hydra, it is straightforward to reproduce the results in section \ref{experiments}. For example to reproduce the fixed-budget and any-time performance results of SGBS+EAS on the \emph{XE\textsubscript{1}} benchmark set, the three steps to follow would be: 
\begin{enumerate}
    \item \texttt{python run\_SGBS.py policy=sgbs\_EAS env=cvrp\_XE\_uch XE\_type=XE\_1 test\_cfg.time\_limit=implicit number\_runs=3 test\_cfg.save\_for\_analysis=True}
    \item \texttt{from models.analyse import average\_run\_results}
    \item \texttt{avg\_res = average\_run\_results(} \\
    \texttt{path\_to\_results="outputs/saved\_results/XE/XE\_1/TL\_implicit", model\_name="SGBS-EAS", number\_runs=3)} \# gives dict with summary stats
\end{enumerate}

The same three steps apply to other methods and algorithms by changing the \texttt{model}-name in \texttt{run\_SGBS.py} to the respective method or other datasets by changing the \texttt{env} argument to another data-type or benchmark set environment, such as \texttt{cvrp\_100\_unf} or \texttt{cvrp\_X\_uch}. Note that the argument \texttt{XE\_type} is unique to the \texttt{cvrp\_XE\_uch} environment. The value \texttt{implicit} for the time limit returns a time limit according to the problem size ($T_{\scaleto{\text{MAX}}{3.5pt}} = 2.4N$). Also, for demonstration and testing purposes, one can set the argument \texttt{test\_cfg.dataset\_size=2} to get a quick check on the functionality and workings of a method.

\subsection{Existing Baselines}
\label{app:baselines}
The suite comes ready with a set of already integrated baselines from relevant publications, mostly based on their available source code. In the following we will give a brief description for each of them.

\paragraph{Savings-CW}
The savings algorithm \cite{savings} is a simple but relatively effective constructive heuristic for the CVRP. It starts by constructing an initial feasible solution where each customer is assigned a separate route at first. Afterwards, the routes are sequentially merged based on the distance that is saved by the merge until no feasible merge can be performed anymore. We utilize the open source implementation\footnote{https://github.com/yorak/VeRyPy} from \cite{rasku2019meta} which is licensed under the MIT license. 

\paragraph{LKH3}
LKH3 is a powerful meta-heuristic for solving a variety of related routing problems \cite{helsgaun2017extension}. In its core it is based on an iterative k-opt local search which is enhanced in several ways, such as for example graph sparsification (candidate set generation), and solution recombination to improve solution quality and computational efficiency. It was originally devised for the TSP but now includes extensions to many problem types by minimizing penalty functions to deal with constraints or transforming the problem to an equivalent TSP if such one exists. The source code\footnote{http://webhotel4.ruc.dk/~keld/research/LKH-3/} is available freely for academic and non-commercial use only.

\paragraph{ORT}
The Google OR-Tools \cite{ortools} is a general software framework to tackle various problems from Operations Research, such as Vehicle Routing, Scheduling, Packing, etc. It also comes with an interface for solving Linear Programming, Constraint Programming and Mixed Integer Programming with multiple options for backbone solvers. The routing interface allows to flexibly design routing problems. We include an ORT implementation for the CVRP. For the experimental results we utilize two settings: (i) a large neighborhood search with a Simulation Annealing acceptance criterion (just called SA in the paper) and a Guided Local Search (GLS). The code\footnote{https://github.com/google/or-tools} is licensed under the Apache Software License (Apache 2.0)

\paragraph{MDAM}
The Multi-Decoder Attention Model \cite{kool2018attention} is an extension of the Attention Model \cite{kool2018attention}, a self-attention based model for solving routing problems trained with reinforcement learning. MDAM extends the model by training with multiple decoder heads regularized with an additional KL-Divergence loss to promote diversity among the multiple decoders. At inference time, each of the heads can be decoded separately either with a greedy rollout or a per head beam search. The code\footnote{https://github.com/liangxinedu/MDAM} is available under the MIT license.

\paragraph{NLNS}
Neural Large Neighborhood search \cite{hottung2019neural} is a machine learning based variant of a large neighborhood search defined by a set of heuristic destroy operators, destroying parts of a solution at random and by a set of learnable repair operators. The repair operators are parameterized by a neural network that repeatedly connects end nodes of two incomplete routes to each other or to the depot until a destroyed solution is complete. The method can be either run in "single-instance" or "batch-mode", where "single-instance" always keeps a set of incumbent solutions that get destroyed and repaired solutions in parallel while solving a single instance, while "batch-mode" rather parallelizes over multiple instances in a batch. We use here the "single-instance" setting since it reflects our evaluation protocol. We assume that in a typical application one has to solve one problem at a time at a given time budget rather than having to continuously solve many problems in parallel. The code is open source\footnote{https://github.com/ahottung/NLNS} and licensed under the GNU General Public License v3.0.

\paragraph{POMO}
Policy Optimization with Multiple Optima \cite{kwon2020pomo} proposed a new training and inference mechanism for constructive models. They adjust the baseline function in the policy gradient, averaging over multiple rollouts with different starting nodes for a problem instance to get a better baseline estimate. For the TSP, this leverages the solution symmetry of the optimal solution, since the optimal solution can be equivalently constructed in multiple ways if solutions are sequentially constructed as any starting node is valid. For the CVRP, one needs to start at the depot and not all nodes afterwards can still guarantee optimality, however \cite{kwon2020pomo} show that it still useful for the CVRP in practice. They additionally introduce a related inference mechanism where instead of sampling or doing a beam search, a greedy inference over all possible starting positions is done. The code \footnote{https://github.com/yd-kwon/POMO} is available under the MIT license.

\paragraph{DACT}
The Dual Aspect Collaborative Transformer \cite{ma2021learning} is an approach of a learned local search, where the models action space is not selecting the local search operator to use but selecting a singular next solution from a neighborhood function. The 2-opt neighborhood for instance is parameterized in this way by selecting a pair of nodes. The used architecture includes two separate transformer encoders for the static problem and current solution with a cyclical positional encoding and the decoder combines the information from both to predict the action. The code\footnote{https://github.com/yining043/VRP-DACT} is open source under the MIT license.

\paragraph{FILO}
FILO \cite{accorsi2021fast} is a state-of-the-art pure iterated local search approach for solving large-scale CVRPs. It includes many techniques to keep the optimization procedure efficient and localized but also effective by building up a large set of neighborhood functions. The code\footnote{https://github.com/acco93/filo} is open source under the GNU Affero General Public License v3.0

\paragraph{NeuroLKH}
NeuroLKH \cite{xin2021neurolkh} is a machine learning based variant of LKH3 \cite{helsgaun2017extension} (see above). When solving an instance, LKH3 first invokes a candidate set generation procedure that removes the edges from the graph that according to some criterion are not likely to be contained in the optimal solution and thus not considered anymore. This essentially sparsifies the graph making the subsequent local search more efficient. NeuroLKH trains a supervised graph neural network from past solutions to replace the heuristic procedure that emits the candidate set. The code is open source\footnote{https://github.com/liangxinedu/NeuroLKH} with no specific license attached. 

\paragraph{SGBS}
Simulation Guided Beam Search \cite{choo2022simulation} is an improved inference strategy for constructive models, which often utilize simple post-hoc tree searches such as a greedy rollout, sampling or a beam search. SGBS evaluates the candidates in the beam more carefully by not only pruning based on the joint probability of the model on the partial solution but also on the actual score based on a rollout. The procedure is further hybridized with efficient active search (EAS) \cite{eas} which finetunes the model during the inference procedure. The code\footnote{https://github.com/yd-kwon/sgbs} is open source under the MIT license.

\paragraph{HGS}
HGS-CVRP is a state-of-the-art OR meta heuristic, originally proposed in 2012 by \citet{vidal2012hybrid}. An updated variant of it, specialized to the CVRP was published in 2022 by \citet{vidal2022hybrid} alongside an open source implementation. The code\footnote{https://github.com/vidalt/HGS-CVRP} is licensed under the MIT license. HGS is a hybrid strategy of a genetic algorithm and a local search. In brief, it creates a pool of initial solutions and then at each step (i) new solutions are created by recombining solutions from the pool, (ii) these solutions are optimized with a local search, (iii) and the pool is updated with the newly found solutions. For this general strategy to be effective, the pool needs to be precisely controlled with respect to size, solution quality, diversity and a highly effective local search is needed, since this is still the main component for improving solutions. For more details on these strategies, see \cite{vidal2012hybrid, vidal2022hybrid}. 

\paragraph{NeuroLS}
NeuroLS \cite{falkner2023learning} trains a meta controller parameterized by a GNN that controls the higher level strategies of an iterated local search procedure. The actions include controlling whether the solution should be accepted, which LS operator to choose next and whether the solution should be perturbed. The code\footnote{https://github.com/jokofa/NeuroLS} is open source with no specific license provided.

\section{Experimental Details}
\label{app:expDetails}
The complete configuration of the models for the experimental setup can be found in the methods' \texttt{config} directories in the Routing Arena repository. Here we want to highlight the experimental details concerning the methods evaluated in section \ref{experiments}. 

\paragraph{Data.} The 128 \emph{Unif100} instances for the experiments in this work have been sampled according the uniform distribution in \cite{kool2018attention}. The coordinates of the instances lie between 0 and 1, and the uniform demands sampled between 1 and 10 have been normalized by the original capacity (50), such that the the normalized vehicle capacity during the inference is 1.0. This input works as is for most of the ML-based methods, however most of the OR-based methods, such as ORT \cite{ortools} expect integer values as inputs. For that reason and for those methods, we multiply the uniform coordinates, demands and capacities with a integer precision of 10000 for the execution of the inference, while for the evaluation the values are scaled back. Concerning the \emph{XML} and the \emph{XE} data, the process is reversed as the coordinates of the instances are already integers and lie on a grid size of 1-1000 and the demands are integer as well. The data is normalized before inference for most ML-based methods, while it is used in its original form for the OR methods.

\paragraph{Experimental Setup.} The experimental runs for the Fixed Budget Evaluation and the Any-time Evaluation are conducted simultaneously, where we give the methods a maximal time budget of $T_{\scaleto{\text{MAX}}{3.5pt}}=120$ seconds for the \emph{Unif100} dataset, a limit of $T_{\scaleto{\text{MAX}}{3.5pt}}=240$ seconds for the \emph{XML} dataset and a runtime budget of $T_{\scaleto{\text{MAX}}{3.5pt}}=2.4N$ seconds for the \emph{XE} datasets. We justify a smaller time budget of 120 seconds (instead of the implicit, size-dependent budget of 240) for uniformly distributed dataset with the fact that these instances are in comparison to \emph{X}-type distributed data, not as challenging to solve. Moreover, we see from Fig. \ref{fig:trajectories_unif100} that the trajectories of most methods seem to be "converged" in the sense that not a lot of improvements are seen after 40\% of the time limit (48 seconds), except for NeuroLS.

\paragraph{Models.} For the detailed lists of the hyperparameters used to conduct the evaluations, we refer the reader to the configuration files of the respective models in the provided code. For most parts they are not changed. Generally we want to mention that, due to time constraints, we have not re-tuned all hyperparameters of all ML models for all the different datasets. We have however attempted to retrain many models concerning the more challenging \emph{X}-distribution, where some models (for example MDAM \cite{xin2021multi} and NeuroLS \cite{falkner2023learning}) still performed better using the uniform-data distribution model-checkpoint provided by the authors. Even though the number of epochs and amount of training data sampled was the same, we assume that more training time may would have been needed for those models to learn the \emph{X}-distribution and we aim to improve on this aspect in the next experiments.
The method configurations for the experiments in section \ref{experiments} are as follows: For \textbf{POMO+SA} we have mostly used the default setting, consisting of the sampling-strategy for the rollout (instead of the greedy rollout) and an augmentation factor of 8, except for the \emph{XE}-sets for which the greedy-rollout results are displayed. We will add the sampling results in the next revision. Concerning the performed LNS by ORT, we have considered some additional search parameters that are not included in the default version of the ORT search, such as relocation- and cross-exchange- operators, which are detailed in the respective \texttt{config} file. For \textbf{NLNS}, we have opted for the single-instance setting instead of the batch inference, firstly because a per-instance inference scheme is employed for all methods evaluated and the batched inference setting of the model is not designed for per-instance evaluation and also not trivial to change in the code. Regarding \textbf{Neuro-LKH}, the results are shown for the default (originally published) version. Note that we have not re-trained the model as this would have required to generate a large amount of training data first, albeit it shows competitive performances. For \textbf{SGBS+EAS}, the default sampling-rollout strategy was selected with an additional EAS, the number of SGBS-EAS iterations has been kept at the default value of 28. The settings for \textbf{NeuroLS} have been kept as in the original source code, we have increased the number of iterations in some larger instances cases, where the time limit is also increased. For the \textbf{Savings+SA} implementations we have

\section{Background on Primal Integral}
\label{app:pi}



The PI metric presented in Equation \ref{PI_dimacs} is based on the primal gap function, $p(t)$, and defined as a performance measure in \cite{berthold2013measuring} to evaluate the solution quality development over the course of the optimization process. The idea of the PI is to have a more tailored performance criteria for primal heuristics, which have as a main target the optimal "trade-off between speed and solution quality" (\cite{berthold2013measuring}).
This section briefly introduces the background for the normalized Primal Integral measure presented in section \ref{metrics} and refer the reader to \citet{berthold2013measuring} for detailed explanations.

 Given a solution time budget $T_{\scaleto{\text{MAX}}{3.5pt}} \in \R_{\geq 0}$, the primal gap function $p: [0, T_{\scaleto{\text{MAX}}{3.5pt}}] \rightarrow [0,1]$ is defined as

\begin{equation}
\begin{aligned}
\label{primal_gap_step_func}
p(t) := 
     \begin{cases}
         1, & \text{if} \quad \text{no incumbent solution so far},\\
         \gamma(\tilde{x}(t)), & \text{incumbent solution } \tilde{x} \text{ at time } t\\
     \end{cases}   
\end{aligned}
\end{equation}
where $\gamma(\tilde{x})$ is the primal gap
\begin{equation}
\begin{aligned}
\label{app:primal_gap}
\gamma(\tilde{x}) := 
     \begin{cases}
         0, & \text{if} \quad |c^T\tilde{x}_{\text{opt}}| = |c^T\tilde{x}| = 0,\\
         1, & \text{if} \quad |c^T\tilde{x}_{\text{opt}}| \cdot |c^T\tilde{x}| < 0,\\
     \frac{|c^T\tilde{x}_{\text{opt}} \ - \ c^T\tilde{x}|}{\max \{ |c^T\tilde{x}_{\text{opt}}, c^T\tilde{x}| \} } & \text{else}\\
     \end{cases}   
\end{aligned}
\end{equation}

The Primal Integral $P(T)$ for the last found solution of a solver at time $T \in [0, T_{\scaleto{\text{MAX}}{3.5pt}}]$ then essentially is the integral of the primal gap criteria of incumbent (i.e. intermediate) solutions normalized by the intermediate run-times between the recordings of incumbent solutions $i \in 1,\ldots,n-1$ given that $t_0 = 0$, $t_i \in [0,T]$ and $t_n=T$:

\phantom{a}
\begin{equation}
\label{app:normed_PI}
         \hspace{0.1 cm}
          P(T) = \int_{t=0}^{T} p(t) \phantom{l}dt = \sum^{n}_{i=1}p(t_{i-1})\cdot (t_i - t_{i-1})
 \end{equation}
 \citet{berthold2013measuring} proposes to use $P(T_{\scaleto{\text{MAX}}{3.5pt}})$ as a measure for the quality of a primal heuristic, where smaller values are better. 
 The version of this metric used in the 12th DIMACS Challenge on Vehicle Routing is a scaled, normalized version of this originally defined primal integral in Equation \ref{app:normed_PI}, that essentially captures the average solution quality ($P(T_{\scaleto{\text{MAX}}{3.5pt}})/T_{\scaleto{\text{MAX}}{3.5pt}}$).


\section{Further Experimental Results}
\label{app:furtherResults}
This section constitutes on one hand a more detailed view on the experiments in section \ref{experiments}, including the standard deviations over the three runs, and on the other hand further experiments on the two other datasets highlighted in Table \ref{overview} that are currently not included in the "Benchmark-Test", but nevertheless provide interesting insights to the performance ranking.

\begin{table}[htb]
  \caption{Relative Gap (in \%) $\pm$ std. dev. for a selection of methods and datasets. The results are averaged over three runs for a time budget in seconds of $T_{\scaleto{\text{MAX}}{3.5pt}} = 2.4N$}
  \label{table:gap2}
  \centering
  \setlength\tabcolsep{2.5pt}
    \resizebox{\columnwidth}{!}{
    \renewcommand{\arraystretch}{1.5} 
  \begin{tabular}{l||l|l|l|l|l||l|l|l|l}  
    \toprule
    \textbf{Gap (\%)} & \multicolumn{5}{c||}{ML} & \multicolumn{4}{c}{OR}  \\
    Dataset     & \multicolumn{1}{c}{POMO+SA}              & \multicolumn{1}{c}{NLNS}              & \multicolumn{1}{c}{NeuroLKH}          & \multicolumn{1}{c}{SGBS+EAS}              & \multicolumn{1}{c||}{NLS}               & \multicolumn{1}{c}{Savings+SA}           & \multicolumn{1}{c}{LKH}               & \multicolumn{1}{c}{FILO}              & \multicolumn{1}{c}{HGS}               \\
    \midrule
    Unif100                  & 1.5755 $\pm$ 0.0035  & 188.38 $\pm$ 0.1830   & 1.0910 $\pm$ 0.0004   & \textit{\underline{0.2188  $\pm$ 0.0065}}  & 2.4092 $\pm$ 0.0642 & 4.0853 $\pm$ 0.0000 & 1.1387 $\pm$ 0.0004 & 0.9546 $\pm$ 0.0105 & \textbf{0.0058 $\pm$ 0.0000} \\
    XML                      & 3.5290 $\pm$ 0.0000  & 164.08 $\pm$ 0.4707   & \textit{0.2261 $\pm$ 0.0003}  & 112.57  $\pm$ 0.0301 & 3.2930 $\pm$ 0.0510 & 5.4750 $\pm$ 0.0000 & 0.1848 $\pm$ 0.0000 & \underline{0.0612 $\pm$ 0.0017} & \textbf{0.0258 $\pm$ 0.0000} \\
    XE\textsubscript{1}      & 1.7899 $\pm$ 0.0000  & 1.6509 $\pm$ 0.0709   & \textit{0.5375 $\pm$ 0.0000}  & 1.2264  $\pm$ 0.0179 & 1.0742 $\pm$ 0.0258 & 3.4786 $\pm$ 0.0000 & 0.4622 $\pm$ 0.0000 & \underline{0.1600 $\pm$ 0.0163} & \textbf{0.0007 $\pm$ 0.0000} \\
    XE\textsubscript{2}      & 2.4171 $\pm$ 0.0000  & 3.0051 $\pm$ 0.1570   & \textit{1.0733 $\pm$ 0.0029}  & 2.1091  $\pm$ 0.0313 & 1.4304 $\pm$ 0.0317 & 4.9085 $\pm$ 0.0194 & 0.9621 $\pm$ 0.0000 & \underline{0.3143 $\pm$ 0.0573} & \textbf{0.0013 $\pm$ 0.0003} \\
    XE\textsubscript{3}      & 1.6347 $\pm$ 0.0000  & 2.9922 $\pm$ 0.1282   & \textit{0.4989 $\pm$ 0.0136}  & 0.6543  $\pm$ 0.0098 & 3.2823 $\pm$ 0.0251 & 2.6362 $\pm$ 0.0000 & 0.4170 $\pm$ 0.0000 & \underline{0.0868 $\pm$ 0.0064} & \textbf{0.0045 $\pm$ 0.0008} \\
    XE\textsubscript{4}      & 2.5644 $\pm$ 0.0000  & 4.0368 $\pm$ 0.2021   & \textit{0.2577 $\pm$ 0.0000}  & 1.2007  $\pm$ 0.0563 & 2.8620 $\pm$ 0.0360 & 6.8782 $\pm$ 0.0024 & 0.2918 $\pm$ 0.0000 & \underline{0.0650 $\pm$ 0.0059} & \textbf{0.0016 $\pm$ 0.0000} \\
    XE\textsubscript{5}      & 1.5988 $\pm$ 0.0000  & 3.5959 $\pm$ 0.0763   & \textit{0.0447 $\pm$ 0.0000}  & 1.5290  $\pm$ 0.0230 & 1.1034 $\pm$ 0.0075 & 2.0178 $\pm$ 0.0002 & 0.0460 $\pm$ 0.0000 & \underline{0.0037 $\pm$ 0.0007} & \textbf{0.0034 $\pm$ 0.0000} \\
    XE\textsubscript{6}      & 3.9067 $\pm$ 0.0001  & 7.8519 $\pm$ 0.0720   & \textit{0.4863 $\pm$ 0.0000}  & 2.0510  $\pm$ 0.0460 & 3.5495 $\pm$ 0.0269 & 6.1936 $\pm$ 0.0000 & 0.4687 $\pm$ 0.0000 & \underline{0.1492 $\pm$ 0.0238} & \textbf{0.0051 $\pm$ 0.0025} \\
    XE\textsubscript{7}      & 2.9761 $\pm$ 0.0006  & 6.9155 $\pm$ 0.0851   & \textit{0.7493 $\pm$ 0.0007}  & 2.2720  $\pm$ 0.0400 & 1.7228 $\pm$ 0.0307 & 4.4406 $\pm$ 0.0051 & 0.7518 $\pm$ 0.0002 & \underline{0.1758 $\pm$ 0.0289} & \textbf{0.0002 $\pm$ 0.0001} \\
    XE\textsubscript{8}      & 4.2016 $\pm$ 0.0043  & 6.2924 $\pm$ 0.2111   & \textit{0.5065 $\pm$ 0.0000}  & 2.7357  $\pm$ 0.0124 & 3.0172 $\pm$ 0.0531 & 5.7614 $\pm$ 0.0205 & 0.3896 $\pm$ 0.0000 & \underline{0.0760 $\pm$ 0.0174} & \textbf{0.0005 $\pm$ 0.0000} \\
    XE\textsubscript{9}      & 4.3875 $\pm$ 0.0000  & 9.2882 $\pm$ 0.4064   & \textit{0.7622 $\pm$ 0.0000}  & 3.6175  $\pm$ 0.0157 & 3.3387 $\pm$ 0.0412 & 5.8388 $\pm$ 0.0047 & 0.6636 $\pm$ 0.0000 & \underline{0.0552 $\pm$ 0.0099} & \textbf{0.0092 $\pm$ 0.0001} \\
    XE\textsubscript{10}     & 0.5005 $\pm$ 0.0000  & 0.1962 $\pm$ 0.0104   & \textit{0.0082 $\pm$ 0.0000}  & 1.1656  $\pm$ 0.0417 & 46.978 $\pm$ 0.0000 & 0.4093 $\pm$ 0.0003 & 0.0084 $\pm$ 0.0000 & \underline{0.0017 $\pm$ 0.0007} & \textbf{0.0005 $\pm$ 0.0000} \\
    XE\textsubscript{11}     & 4.1613 $\pm$ 0.0018  & 4.4449 $\pm$ 0.0496   & \textit{0.2994 $\pm$ 0.0000}  & 5.0140  $\pm$ 0.0331 & 8.7922 $\pm$ 0.0686 & 5.9679 $\pm$ 0.0032 & 0.4322 $\pm$ 0.0000 & \underline{0.0514 $\pm$ 0.0077} & \textbf{0.0005 $\pm$ 0.0001} \\
    XE\textsubscript{12}     & 2.8668 $\pm$ 0.0005  & 4.5255 $\pm$ 0.0925   & \textit{0.7907 $\pm$ 0.0020}  & 3.1513  $\pm$ 0.0827 & 4.1838 $\pm$ 0.0287 & 2.1960 $\pm$ 0.0034 & 0.7880 $\pm$ 0.0003 & \underline{0.1295 $\pm$ 0.0099} & \textbf{0.0094 $\pm$ 0.0029} \\
    XE\textsubscript{13}     & 6.0626 $\pm$ 0.0000  & 6.7343 $\pm$ 0.0947   & \textit{1.1545 $\pm$ 0.0032}  & 7.5734  $\pm$ 0.1329 & 2.6117 $\pm$ 0.0446 & 4.5258 $\pm$ 0.0000 & 0.7490 $\pm$ 0.0000 & \underline{0.1793 $\pm$ 0.0097} & \textbf{0.0000 $\pm$ 0.0000} \\
    XE\textsubscript{14}     & 3.8479 $\pm$ 0.0001  & 5.4175 $\pm$ 0.1045   & \textit{0.4165 $\pm$ 0.0000}  & 4.4713  $\pm$ 0.0610 & 10.966 $\pm$ 0.0197 & 2.5789 $\pm$ 0.0017 & 0.1943 $\pm$ 0.0000 & \underline{0.0153 $\pm$ 0.0062} & \textbf{0.0053 $\pm$ 0.0000} \\
    XE\textsubscript{15}     & 5.7831 $\pm$ 0.0011  & 7.2811 $\pm$ 0.2195   & \textit{1.6926 $\pm$ 0.0021}  & 6.8795  $\pm$ 0.0435 & 3.8426 $\pm$ 0.0326 & 4.2942 $\pm$ 0.0000 & 1.1491 $\pm$ 0.0008 & \underline{0.4020 $\pm$ 0.0255} & \textbf{0.0001 $\pm$ 0.0001} \\
    XE\textsubscript{16}     & 6.9635 $\pm$ 0.0000  & 4.3552 $\pm$ 0.1003   & \textit{1.1686 $\pm$ 0.0027}  & 7.9388  $\pm$ 0.1495 & 2.1257 $\pm$ 0.0351 & 3.4119 $\pm$ 0.0000 & 0.9413 $\pm$ 0.0000 & \underline{0.2119 $\pm$ 0.0157} & \textbf{0.0008 $\pm$ 0.0000} \\
    XE\textsubscript{17}     & 7.9297 $\pm$ 0.0052  & 8.8609 $\pm$ 0.0121   & \textit{1.0255 $\pm$ 0.0000}  & 6.0329  $\pm$ 0.0331 & 3.8086 $\pm$ 0.0551 & 4.2504 $\pm$ 0.0021 & 1.0416 $\pm$ 0.0000 & \underline{0.2385 $\pm$ 0.0127} & \textbf{0.0019 $\pm$ 0.0006} \\
    \bottomrule
    AVG                      & 3.6156 $\pm$	0.0009  & 23.153 $\pm$ 0.1445   & \textit{0.6731 $\pm$ 0.0015}	& 9.0745  $\pm$	0.0456 & 5.8100 $\pm$ 0.0357 & 4.1762 $\pm$	0.0033 & 0.5832 $\pm$ 0.0000 & \underline{0.1753 $\pm$ 0.0140} & \textbf{0.0040 $\pm$ 0.0004} \\
    \bottomrule
  \end{tabular} }
\end{table}

Looking at the average standard deviation across the benchmark-test datasets, we see that the methods overall provide robust results. NLNS exhibits a higher standard deviation, which is mainly due to the inferior results on the XML benchmark set.

\begin{table}[htb]
  \caption{Primal Integral (PI) for a selection of methods and datasets. The results are averaged over three runs for a time budget in seconds of $T_{\scaleto{\text{MAX}}{3.5pt}} = 2.4N$}
  \label{table:pi2}
  \centering
  \setlength\tabcolsep{2.5pt}
    \resizebox{\columnwidth}{!}{
    \renewcommand{\arraystretch}{1.5} 
  \begin{tabular}{l||l|l|l|l|l||l|l|l|l}  
    \toprule
    \textbf{PI} & \multicolumn{5}{c||}{ML} & \multicolumn{4}{c}{OR}  \\
    Dataset     & \multicolumn{1}{c}{POMO+SA}              & \multicolumn{1}{c}{NLNS}              & \multicolumn{1}{c}{NeuroLKH}          & \multicolumn{1}{c}{SGBS+EAS}              & \multicolumn{1}{c||}{NLS}               & \multicolumn{1}{c}{Savings+SA}           & \multicolumn{1}{c}{LKH}               & \multicolumn{1}{c}{FILO}              & \multicolumn{1}{c}{HGS}               \\
    \midrule
    Unif100                    & 1.6242 $\pm$ 0.0021 & 10.000 $\pm$ 0.0000 & 1.3742 $\pm$ 0.0004 & \textit{\underline{0.5708 $\pm$ 0.0032}} & 2.8658 $\pm$ 0.0387 & 4.1387 $\pm$ 0.0001 & 1.3620 $\pm$ 0.0001 & 1.0403 $\pm$ 0.0102 & \textbf{0.0124 $\pm$ 0.0000}  \\
    XML                        & 2.9444 $\pm$ 0.0001 & 10.000 $\pm$ 0.0000 & \textit{0.3671 $\pm$ 0.0004} & 9.0448 $\pm$ 0.0183 & 2.9920 $\pm$ 0.0000 & 4.5515 $\pm$ 0.0003 & 0.2667 $\pm$ 0.0001 & \underline{0.1100 $\pm$ 0.0022} & \textbf{0.0300 $\pm$ 0.0000}  \\
    XE\textsubscript{1}        & 1.8390 $\pm$ 0.0005 & 5.1683 $\pm$ 0.0589 & \textit{0.8175 $\pm$ 0.0031} & 2.0927 $\pm$ 0.0211 & 1.3083 $\pm$ 0.0128 & 3.5038 $\pm$ 0.0005 & 0.6280 $\pm$ 0.0002 & \underline{0.2208 $\pm$ 0.0116} & \textbf{0.0065 $\pm$ 0.0000}  \\
    XE\textsubscript{2}        & 2.4789 $\pm$ 0.0004 & 6.0415 $\pm$ 0.0211 & \textit{1.3860 $\pm$ 0.0056} & 3.0812 $\pm$ 0.0510 & 1.7499 $\pm$ 0.0326 & 5.3230 $\pm$ 0.0041 & 1.1981 $\pm$ 0.0006 & \underline{0.4298 $\pm$ 0.0356} & \textbf{0.0106 $\pm$ 0.0002}  \\
    XE\textsubscript{3}        & 1.6941 $\pm$ 0.0002 & 6.1592 $\pm$ 0.0547 & \textit{0.7877 $\pm$ 0.0073} & 1.0085 $\pm$ 0.0255 & 3.5035 $\pm$ 0.0218 & 2.7585 $\pm$ 0.0011 & 0.6550 $\pm$ 0.0006 & \underline{0.1320 $\pm$ 0.0050} & \textbf{0.0268 $\pm$ 0.0007}  \\
    XE\textsubscript{4}        & 2.6090 $\pm$ 0.0003 & 6.9602 $\pm$ 0.1287 & \textit{0.3684 $\pm$ 0.0004} & 2.0927 $\pm$ 0.0391 & 3.1931 $\pm$ 0.0496 & 7.0084 $\pm$ 0.0012 & 0.3880 $\pm$ 0.0003 & \underline{0.0886 $\pm$ 0.0027} & \textbf{0.0117 $\pm$ 0.0001}  \\
    XE\textsubscript{5}        & 1.6885 $\pm$ 0.0003 & 7.5302 $\pm$ 0.0919 & \textit{0.1012 $\pm$ 0.0006} & 2.8230 $\pm$ 0.0230 & 1.2211 $\pm$ 0.0117 & 2.2221 $\pm$ 0.0023 & 0.0793 $\pm$ 0.0001 & \underline{0.0219 $\pm$ 0.0006} & \textbf{0.0055 $\pm$ 0.0000}  \\
    XE\textsubscript{6}        & 4.0104 $\pm$ 0.0005 & 9.7696 $\pm$ 0.0733 & \textit{0.7374 $\pm$ 0.0004} & 3.5330 $\pm$ 0.0530 & 3.9051 $\pm$ 0.0616 & 6.3674 $\pm$ 0.0030 & 0.6072 $\pm$ 0.0003 & \underline{0.2027 $\pm$ 0.0214} & \textbf{0.0207 $\pm$ 0.0005}  \\
    XE\textsubscript{7}        & 3.1032 $\pm$ 0.0004 & 9.0656 $\pm$ 0.0689 & \textit{1.0097 $\pm$ 0.0024} & 3.3500 $\pm$ 0.0240 & 1.9510 $\pm$ 0.0116 & 4.5930 $\pm$ 0.0023 & 0.9326 $\pm$ 0.0006 & \underline{0.2801 $\pm$ 0.0190} & \textbf{0.0171 $\pm$ 0.0001}  \\
    XE\textsubscript{8}        & 4.2816 $\pm$ 0.0044 & 9.1446 $\pm$ 0.0679 & \textit{0.7455 $\pm$ 0.0028} & 3.7137 $\pm$ 0.0289 & 3.2953 $\pm$ 0.0632 & 5.9378 $\pm$ 0.0019 & 0.5144 $\pm$ 0.0001 & \underline{0.1475 $\pm$ 0.0120} & \textbf{0.0094 $\pm$ 0.0001}  \\
    XE\textsubscript{9}        & 4.3995 $\pm$ 0.0002 & 9.7343 $\pm$ 0.0764 & \textit{0.9567 $\pm$ 0.0006} & 4.7665 $\pm$ 0.0642 & 3.6502 $\pm$ 0.0238 & 5.9425 $\pm$ 0.0015 & 0.7552 $\pm$ 0.0004 & \underline{0.1457 $\pm$ 0.0125} & \textbf{0.0248 $\pm$ 0.0001}  \\
    XE\textsubscript{10}       & 0.5299 $\pm$ 0.0009 & 2.9246 $\pm$ 0.0048 & \textit{0.0629 $\pm$ 0.0024} & 1.8562 $\pm$ 0.0168 & 10.000 $\pm$ 0.0000 & 0.4347 $\pm$ 0.0001 & 0.0524 $\pm$ 0.0004 & \underline{0.0051 $\pm$ 0.0007} & \textbf{0.0031 $\pm$ 0.0001}  \\
    XE\textsubscript{11}       & 4.2309 $\pm$ 0.0003 & 7.2121 $\pm$ 0.0705 & \textit{0.4783 $\pm$ 0.0019} & 7.0477 $\pm$ 0.0682 & 8.9877 $\pm$ 0.0721 & 6.1452 $\pm$ 0.0010 & 0.5751 $\pm$ 0.0002 & \underline{0.1076 $\pm$ 0.0073} & \textbf{0.0269 $\pm$ 0.0007}  \\
    XE\textsubscript{12}       & 2.9299 $\pm$ 0.0004 & 8.0742 $\pm$ 0.0491 & \textit{1.0292 $\pm$ 0.0019} & 4.2689 $\pm$ 0.0432 & 4.3974 $\pm$ 0.0030 & 2.2253 $\pm$ 0.0002 & 1.0216 $\pm$ 0.0009 & \underline{0.2056 $\pm$ 0.0083} & \textbf{0.0876 $\pm$ 0.0041}  \\
    XE\textsubscript{13}       & 6.1788 $\pm$ 0.0007 & 9.1827 $\pm$ 0.0453 & \textit{1.6167 $\pm$ 0.0020} & 8.8134 $\pm$ 0.0433 & 2.9676 $\pm$ 0.0192 & 4.5782 $\pm$ 0.0004 & 0.9769 $\pm$ 0.0006 & \underline{0.2931 $\pm$ 0.0117} & \textbf{0.0294 $\pm$ 0.0020}  \\
    XE\textsubscript{14}       & 3.9250 $\pm$ 0.0006 & 7.3287 $\pm$ 0.0710 & \textit{0.6732 $\pm$ 0.0019} & 5.7794 $\pm$ 0.1315 & 9.9980 $\pm$ 0.0026 & 2.6188 $\pm$ 0.0002 & 0.2590 $\pm$ 0.0004 & \underline{0.0715 $\pm$ 0.0064} & \textbf{0.0189 $\pm$ 0.0004}  \\
    XE\textsubscript{15}       & 5.8212 $\pm$ 0.0004 & 8.9290 $\pm$ 0.0843 & \textit{2.1087 $\pm$ 0.0038} & 7.7627 $\pm$ 0.0917 & 4.1188 $\pm$ 0.0210 & 4.3410 $\pm$ 0.0002 & 1.5106 $\pm$ 0.0009 & \underline{0.5571 $\pm$ 0.0202} & \textbf{0.0551 $\pm$ 0.0038}  \\
    XE\textsubscript{16}       & 7.0702 $\pm$ 0.0011 & 6.8002 $\pm$ 0.0365 & \textit{1.5793 $\pm$ 0.0023} & 9.0651 $\pm$ 0.0601 & 2.3633 $\pm$ 0.0514 & 3.4496 $\pm$ 0.0002 & 1.2061 $\pm$ 0.0007 & \underline{0.3330 $\pm$ 0.0067} & \textbf{0.0514 $\pm$ 0.0001}  \\
    XE\textsubscript{17}       & 8.0228 $\pm$ 0.0007 & 9.8576 $\pm$ 0.0254 & \textit{1.2281 $\pm$ 0.0062} & 7.7333 $\pm$ 0.0095 & 4.0653 $\pm$ 0.0413 & 4.2947 $\pm$ 0.0002 & 1.2194 $\pm$ 0.0005 & \underline{0.3809 $\pm$ 0.0095} & \textbf{0.0579 $\pm$ 0.0016}  \\
    \bottomrule
    AVG	                       & 3.6517 $\pm$ 0.0008 & 7.8886 $\pm$ 0.0541 & \textit{0.9173	$\pm$ 0.0024} & 4.6528 $\pm$ 0.0429 & 4.0281	$\pm$ 0.0283 & 4.2334 $\pm$ 0.0011 & 0.7478	$\pm$ 0.0004 & \underline{0.2512 $\pm$ 0.0107} & \textbf{0.0266 $\pm$ 0.0008}  \\
    \bottomrule
  \end{tabular} }
\end{table}

The standard deviations for the PI and WRAP metric are consequently also generally small, but exhibit slightly more variability. Furthermore, the PI metric standard deviations are overall higher compared to the WRAP metric standard deviations. The PI standard deviations show four out of nine methods with values on the second decimal, whereas the WRAP metric standard deviations are all smaller than 0.01.

\begin{table}[htb]
  \caption{Weighted Relative Average Performance (WRAP) for a selection of methods and datasets. The results are averaged over three runs for a time budget in seconds of $T_{\scaleto{\text{MAX}}{3.5pt}} = 2.4N$}
  \label{table:wrap2}
  \centering
  \setlength\tabcolsep{2.5pt}
    \resizebox{\columnwidth}{!}{
    \renewcommand{\arraystretch}{1.5} 
  \begin{tabular}{l||l|l|l|l|l||l|l|l|l}  
    \toprule
    \textbf{WRAP} & \multicolumn{5}{c||}{ML} & \multicolumn{4}{c}{OR}  \\
    Dataset     & \multicolumn{1}{c}{POMO+SA}              & \multicolumn{1}{c}{NLNS}              & \multicolumn{1}{c}{NeuroLKH}          & \multicolumn{1}{c}{SGBS+EAS}              & \multicolumn{1}{c||}{NLS}               & \multicolumn{1}{c}{Savings+SA}           & \multicolumn{1}{c}{LKH}               & \multicolumn{1}{c}{FILO}              & \multicolumn{1}{c}{HGS}               \\
    \midrule
    Unif100                 & 0.3969 $\pm$ 0.0003 & 1.0000 $\pm$ 0.0000 & 0.3465 $\pm$ 0.0002 & \underline{\textit{0.1419 $\pm$ 0.0009}} & 0.6984 $\pm$ 0.0086 & 0.9396 $\pm$ 0.0007 & 0.3509 $\pm$ 0.0000 & 0.2537 $\pm$ 0.0025 & \textbf{0.0028 $\pm$ 0.0000}  \\
    XML                     & 0.5596 $\pm$ 0.0000 & 1.0000 $\pm$ 0.0000 & \textit{\underline{0.0993}} $\pm$ 0.0025 & 0.9954 $\pm$ 0.0009 & 0.5810 $\pm$ 0.0000 & 0.9190 $\pm$ 0.0001 & 0.1309 $\pm$ 0.0007 & \textbf{0.0265 $\pm$ 0.0024} & 0.2911 $\pm$ 0.0049  \\
    XE\textsubscript{1}     & 0.5480 $\pm$ 0.0040 & 0.8414 $\pm$ 0.0148 & \textit{0.2493 $\pm$ 0.0009} & 0.6032 $\pm$ 0.0095 & 0.4012 $\pm$ 0.0018 & 0.9769 $\pm$ 0.0032 & 0.1838 $\pm$ 0.0000 & \underline{0.1201 $\pm$ 0.0172} & \textbf{0.0080 $\pm$ 0.0089}  \\
    XE\textsubscript{2}     & 0.4675 $\pm$ 0.0000 & 0.8258 $\pm$ 0.0021 & \textit{0.2741 $\pm$ 0.0009} & 0.5756 $\pm$ 0.0087 & 0.3515 $\pm$ 0.0045 & 0.9394 $\pm$ 0.0007 & 0.2274 $\pm$ 0.0001 & \underline{0.0910 $\pm$ 0.0048} & \textbf{0.0024 $\pm$ 0.0000}  \\
    XE\textsubscript{3}     & 0.6248 $\pm$ 0.0002 & 0.9917 $\pm$ 0.0040 & \textit{0.2921 $\pm$ 0.0026} & 0.3670 $\pm$ 0.0142 & 0.9820 $\pm$ 0.0023 & 0.9557 $\pm$ 0.0011 & 0.2315 $\pm$ 0.0001 & \underline{0.0563 $\pm$ 0.0063} & \textbf{0.0093 $\pm$ 0.0002}  \\
    XE\textsubscript{4}     & 0.3684 $\pm$ 0.0000 & 0.8502 $\pm$ 0.0154 & \textit{0.0533 $\pm$ 0.0002} & 0.2957 $\pm$ 0.0066 & 0.4546 $\pm$ 0.0074 & 0.9829 $\pm$ 0.0026 & 0.0652 $\pm$ 0.0147 & \underline{0.0119 $\pm$ 0.0005} & \textbf{0.0015 $\pm$ 0.0000}  \\
    XE\textsubscript{5}     & 0.7310 $\pm$ 0.0002 & 0.9986 $\pm$ 0.0012 & \textit{0.0561 $\pm$ 0.0231} & 0.9100 $\pm$ 0.0080 & 0.5572 $\pm$ 0.0053 & 0.9801 $\pm$ 0.0044 & 0.0288 $\pm$ 0.0000 & \underline{0.0086 $\pm$ 0.0003} & \textbf{0.0025 $\pm$ 0.0000}  \\
    XE\textsubscript{6}     & 0.6260 $\pm$ 0.0000 & 0.9999 $\pm$ 0.0001 & \textit{0.1166 $\pm$ 0.0000} & 0.5490 $\pm$ 0.0090 & 0.6132 $\pm$ 0.0087 & 0.9866 $\pm$ 0.0015 & 0.0969 $\pm$ 0.0000 & \underline{0.0332 $\pm$ 0.0038} & \textbf{0.0031 $\pm$ 0.0001}  \\
    XE\textsubscript{7}     & 0.6730 $\pm$ 0.0002 & 0.9986 $\pm$ 0.0013 & \textit{0.2115 $\pm$ 0.0004} & 0.6900 $\pm$ 0.0050 & 0.4328 $\pm$ 0.0024 & 0.9915 $\pm$ 0.0012 & 0.1960 $\pm$ 0.0001 & \underline{0.0694 $\pm$ 0.0045} & \textbf{0.0035 $\pm$ 0.0000}  \\
    XE\textsubscript{8}     & 0.7203 $\pm$ 0.0059 & 0.9931 $\pm$ 0.0043 & \textit{0.1260 $\pm$ 0.0005} & 0.6210 $\pm$ 0.0051 & 0.5557 $\pm$ 0.0096 & 0.9956 $\pm$ 0.0003 & 0.0850 $\pm$ 0.0000 & \underline{0.0269 $\pm$ 0.0045} & \textbf{0.0014 $\pm$ 0.0000}  \\
    XE\textsubscript{9}     & 0.7269 $\pm$ 0.0000 & 1.0000 $\pm$ 0.0000 & \textit{0.1596 $\pm$ 0.0001} & 0.7788 $\pm$ 0.0038 & 0.6277 $\pm$ 0.0025 & 0.9881 $\pm$ 0.0022 & 0.1249 $\pm$ 0.0001 & \underline{0.0278 $\pm$ 0.0021} & \textbf{0.0052 $\pm$ 0.0012}  \\
    XE\textsubscript{10}    & 0.9169 $\pm$ 0.0003 & 0.8438 $\pm$ 0.0069 & \textit{0.0692 $\pm$ 0.0006} & 1.0000 $\pm$ 0.0000 & 1.0000 $\pm$ 0.0000 & 0.9634 $\pm$ 0.0025 & 0.0524 $\pm$ 0.0001 & \underline{0.0104 $\pm$ 0.0016} & \textbf{0.0063 $\pm$ 0.0002}  \\
    XE\textsubscript{11}    & 0.6767 $\pm$ 0.0000 & 0.9227 $\pm$ 0.0086 & \textit{0.0763 $\pm$ 0.0003} & 0.9408 $\pm$ 0.0059 & 1.0000 $\pm$ 0.0000 & 0.9728 $\pm$ 0.0000 & 0.0942 $\pm$ 0.0000 & \underline{0.0181 $\pm$ 0.0006} & \textbf{0.0039 $\pm$ 0.0001}  \\
    XE\textsubscript{12}    & 0.9808 $\pm$ 0.0000 & 1.0000 $\pm$ 0.0000 & \textit{0.4594 $\pm$ 0.0009} & 0.9997 $\pm$ 0.0004 & 1.0000 $\pm$ 0.0000 & 0.9684 $\pm$ 0.0004 & 0.4535 $\pm$ 0.0006 & \underline{0.1295 $\pm$ 0.0102} & \textbf{0.0395 $\pm$ 0.0019}  \\
    XE\textsubscript{13}    & 0.9757 $\pm$ 0.0000 & 1.0000 $\pm$ 0.0000 & \textit{0.3455 $\pm$ 0.0002} & 0.9997 $\pm$ 0.0002 & 0.6513 $\pm$ 0.0036 & 0.9555 $\pm$ 0.0001 & 0.2113 $\pm$ 0.0001 & \underline{0.0725 $\pm$ 0.0033} & \textbf{0.0061 $\pm$ 0.0004}  \\
    XE\textsubscript{14}    & 0.9499 $\pm$ 0.0001 & 1.0000 $\pm$ 0.0000 & \textit{0.2278 $\pm$ 0.0002} & 0.9970 $\pm$ 0.0009 & 1.0000 $\pm$ 0.0000 & 0.9520 $\pm$ 0.0001 & 0.0837 $\pm$ 0.0001 & \underline{0.0323 $\pm$ 0.0027} & \textbf{0.0076 $\pm$ 0.0002}  \\
    XE\textsubscript{15}    & 0.9537 $\pm$ 0.0000 & 0.9998 $\pm$ 0.0001 & \textit{0.4945 $\pm$ 0.0008} & 1.0000 $\pm$ 0.0000 & 0.8674 $\pm$ 0.0064 & 0.9689 $\pm$ 0.0012 & 0.3537 $\pm$ 0.0002 & \underline{0.1803 $\pm$ 0.0066} & \textbf{0.0122 $\pm$ 0.0009}  \\
    XE\textsubscript{16}    & 1.0000 $\pm$ 0.0000 & 0.9998 $\pm$ 0.0001 & \textit{0.4517 $\pm$ 0.0006} & 1.0000 $\pm$ 0.0000 & 0.6959 $\pm$ 0.0136 & 0.9946 $\pm$ 0.0001 & 0.3478 $\pm$ 0.0002 & \underline{0.1656 $\pm$ 0.0246} & \textbf{0.0148 $\pm$ 0.0000}  \\
    XE\textsubscript{17}    & 1.0000 $\pm$ 0.0000 & 1.0000 $\pm$ 0.0000 & \textit{0.3029 $\pm$ 0.0013} & 1.0000 $\pm$ 0.0000 & 0.8900 $\pm$ 0.0063 & 0.9771 $\pm$ 0.0018 & 0.3063 $\pm$ 0.0165 & \underline{0.1346 $\pm$ 0.0134} & \textbf{0.0150 $\pm$ 0.0005}  \\
    \bottomrule
    AVG                     & 0.7314 $\pm$ 0.0006 &	0.9613 $\pm$ 0.0031 & \textit{0.2322 $\pm$ 0.0019} & 0.7613 $\pm$ 0.0042 & 0.7032 $\pm$ 0.0044 & 0.9688 $\pm$ 0.0013 & 0.1907 $\pm$ 0.0018 & \underline{0.0773 $\pm$ 0.0059} & \textbf{0.0230 $\pm$ 0.0010} \\
    \bottomrule
  \end{tabular} }
\end{table}

The smaller values in standard deviations of the WRAP metric demonstrate that WRAP is more suitable for aggregating performances over datasets by capturing real performance signals across experiments.

We present further results on the \citet{golden1998impact} benchmark set and original \citet{uchoa2017new} \emph{X}-set below, that currently are not included in the benchmark-test collection, because they either are fairly outdated and to a great extent optimally solved (\citet{golden1998impact}) or already represented in terms of their distribution (\citet{uchoa2017new}) but still rank among the most popular and recent benchmark sets, respectively, in the OR literature.
Table \ref{tab:gap_GX} - \ref{tab:pi_GX} show the results for the percentage gap, WRAP metric and PI metric values for these benchmark sets.

\begin{table}
  \caption{Percentage Gap for a selection of methods for the \citet{golden1998impact} set and \emph{X}-set in \cite{uchoa2017new}. The results are averaged over three runs for a time budget in seconds of $T_{\scaleto{\text{MAX}}{3.5pt}} = 2.4N$}
  \label{tab:gap_GX}
  \centering
    \resizebox{\columnwidth}{!}{
    \renewcommand{\arraystretch}{1.5} 
  \begin{tabular}{l|lllll|llll}  
    \toprule
    \textbf{Gap (\%)} & \multicolumn{5}{c|}{ML} & \multicolumn{4}{c}{OR}  \\
    Dataset     & POMO+SA              & NLNS              & NeuroLKH          & SGBS+EAS              & NLS               & Savings+SA           & LKH               & FILO              & HGS               \\
    \midrule
    X-set   & 7.3004 $\pm$ 0.0018 & 10.7791 $\pm$ 0.0297 & \textit{1.1012 $\pm$ 0.0008} & -                   & 5.2190 $\pm$ 0.0018 & 4.7797 $\pm$ 0.0392 & 1.178 $\pm$ 0.0000 & \underline{0.3357 $\pm$ 0.0000} & \textbf{0.1457 $\pm$ 0.0003} \\
    Golden  & 11.123 $\pm$ 0.0026 & 173.007 $\pm$ 1.3713 & \textit{2.2434 $\pm$ 0.0000} & 39.935 $\pm$ 0.6505 & 10.461 $\pm$ 0.0053 & 11.682 $\pm$ 0.0043 & 2.757 $\pm$ 0.0000 & \underline{0.8526 $\pm$ 0.0409} & \textbf{0.7799 $\pm$ 0.0030} \\
    \bottomrule
  \end{tabular} }
\end{table}

Overall, the results depict the same picture as for the benchmark-test experiments in \ref{table:gap2} - \ref{table:wrap2}; The best performance in terms of percentage gap, WRAP and PI is delivered by HGS, albeit, especially for the \citet{golden1998impact} set closely followed by FILO. In terms of any-time performance all other methods present PI and WRAP scores that are significantly larger. 

\begin{table}
  \caption{Weighted Relative Average Performance (WRAP) for a selection of methods for the \citet{golden1998impact} set and \emph{X}-set in \cite{uchoa2017new}. The results are averaged over three runs for a time budget in seconds of $T_{\scaleto{\text{MAX}}{3.5pt}} = 2.4N$}
  \label{tab:wrap_GX}
  \centering
    \resizebox{\columnwidth}{!}{
    \renewcommand{\arraystretch}{1.5} 
  \begin{tabular}{l|lllll|llll}  
    \toprule
    \textbf{WRAP} & \multicolumn{5}{c|}{ML} & \multicolumn{4}{c}{OR}  \\
    Dataset     & POMO+SA              & NLNS              & NeuroLKH          & SGBS+EAS              & NLS               & Savings+SA           & LKH               & FILO              & HGS               \\
    \midrule
    X-set   & 0.8857 $\pm$ 0.0001 & 0.9918 $\pm$ 0.0020 & \textit{0.2770 $\pm$ 0.0013} & -                   & 0.7880 $\pm$ 0.0002 & 0.9922 $\pm$ 0.0055 & 0.3006 $\pm$ 0.0000 & \underline{0.1078 $\pm$ 0.0000} & \textbf{0.0554 $\pm$ 0.0001} \\
    Golden  & 0.8869 $\pm$ 0.0003 & 1.0000 $\pm$ 0.0000 & \textit{0.2127 $\pm$ 0.0003} & 0.9966 $\pm$ 0.0029 & 0.8300 $\pm$ 0.0001 & 0.9683 $\pm$ 0.0017 & 0.2715 $\pm$ 0.0002 & \underline{0.0935 $\pm$ 0.0049} & \textbf{0.0864 $\pm$ 0.0000} \\
    \bottomrule
  \end{tabular}}
\end{table}

Concerning the ML-based methods, the ranking for any-time performance on these challenging and larger instances is as follows: NeuroLKH ranks first, followed by NLS, POMO+SA, SGBS+EAS and NLNS. Furthermore, NeuroLKH outperforms LKH on both of the the benchmarks sets.
Notably, the results for SGBS+EAS on the \emph{X}-set are currently missing due to Out-of-Memory issues during inference time on a 48GB VRAM NVIDIA A40 machine. This is due to the fact that the \emph{X} benchmark set features instances with problem sizes up to 1000 and SGBS features a costly sampling procedure, where the sampling size is by default equal to the problem size. We will consider smaller sampling sizes for SGBS in future experiments, but wanted to keep the model parameters consistent for this work's evaluation.

\begin{table}[htb]
  \caption{PI for a selection of methods for the \citet{golden1998impact} set and \emph{X}-set in \cite{uchoa2017new}. The results are averaged over three runs for a time budget in seconds of $T_{\scaleto{\text{MAX}}{3.5pt}} = 2.4N$}
  \label{tab:pi_GX}
  \centering
    \resizebox{\columnwidth}{!}{
    \renewcommand{\arraystretch}{1.5} 
  \begin{tabular}{l|lllll|llll}  
    \toprule
    \textbf{PI} & \multicolumn{5}{c|}{ML} & \multicolumn{4}{c}{OR}  \\
    Dataset     & POMO+SA              & NLNS              & NeuroLKH          & SGBS+EAS              & NLS               & Savings+SA           & LKH               & FILO              & HGS               \\
    \midrule
    X-set   & 6.4610 $\pm$ 0.0004 & 8.7265 $\pm$ 0.0269 & \textit{1.2840 $\pm$ 0.0061} & -                   & 4.7469 $\pm$ 0.0012 & 4.8397 $\pm$ 0.0022 & 1.3943 $\pm$ 0.0001 & \underline{0.4911 $\pm$ 0.0000} & \textbf{0.2356 $\pm$ 0.0003} \\
    Golden  & 9.2338 $\pm$ 0.0004 & 10.000 $\pm$ 0.0000 & \textit{2.2672 $\pm$ 0.0010} & 9.8092 $\pm$ 0.0254 & 9.5286 $\pm$ 0.0010 & 8.8398 $\pm$ 0.0001 & 2.8049 $\pm$ 0.0001 & \underline{0.9118 $\pm$ 0.0403} & \textbf{0.8168 $\pm$ 0.0001} \\
    \bottomrule
  \end{tabular} }
\end{table}

The further results in this section emphasize the results in section \ref{experiments} and show a clear global winner for solving CVRP in a fixed-budget and any-time setting, which is HGS. From the ML perspective, the winner for being the most competitive method in terms of any-time as well as fixed-budget performance is NeuroLKH. This clearly indicates how future Neural routing solvers should be conceptualised to improve on their general competitiveness concerning OR Methods by favoring hybrid ML-OR solution approaches.

\section{Run-Time Standardization}
\label{app:run_time_stdization}

Run-time normalization has an important impact on the comparability of ML- and OR-based methods, especially when it comes to Any-time performance. 
The standardization scheme introduced in section \ref{time_normalization} marks a first step towards comparable run-times and thus a more equitable evaluation for (N)CO methods.
In general the scheme implies that the more resources or more powerful resources a method uses, the more its available run-time ($T_{\scaleto{\text{MAX}}{3.5pt}}$) gets weighted down.
Currently, the Routing Arena employs essentially two normalization schemes, one exclusively for methods that are executed on a single-threaded CPU and another for methods that make use of available GPUs. This is because the PassMark ratings reported for single-thread CPUs and GPU machines have very large discrepancies and thus make it hard to find a common Base Reference that should be used as normalizing factor. 

For the single-thread CPU normalization scheme we adopt the procedure from the 12th DIMACS challenge on Vehicle Routing, where the Base Reference has been set to 2000.
Hence, for a method that runs on a single Intel(R) Core(TM) i7-10850H CPU @ 2.70GHz machine with a passMark rating of 2714, the time limit of 240 seconds is down-scaled to roughly 177 seconds per instance. 

Concerning GPU-usage, instead of using the pure PassMark rating used for CPU run-time calibration, we employ a mixture of CPU and GPU PassMark ratings to calibrate the GPU performed run-times as shown in Formula \ref{GPU_passmark}. This implies that a run executed on a NVIDIA GeForce RTX 3060 machine, with a G3D mark of 17177 and a G2D mark of 979 together with a CPU performance equivalent to the 2714 single CPU mark rating; the adjusted per instance time limit of 240 seconds would then be down-scaled to 207 seconds. 

\section{Broader Impact}
\label{app:BroaderImpact}
We propose a new benchmark for evaluating neural solvers to solve routing problems. This could lead to more efficient and better application specific routing solvers which could improve efficiency for example in delivery systems. We do not see negative ethical or societal implications from our work.

\end{document}